\documentclass[a4paper]{article}
\pdfoutput=1
\usepackage{amsfonts}
\usepackage{hyperref}
\hypersetup{
  pdfinfo={
    Title={Defense Against the Dark Arts},
    Author={Ian Goodfellow},
    Subject={},
    Keywords={}
  }
}
\usepackage{pdfpages}

\author{Ian Goodfellow}
\title{Defense Against the Dark Arts:
An overview of adversarial example security
research and future research directions}

\begin{document}
\maketitle

\begin{abstract}
This article presents a summary of a keynote
lecture at the Deep Learning Security workshop
at IEEE Security and Privacy 2018.
This lecture summarizes the state of the art in
defenses against adversarial examples and provides
recommendations for future research directions on
this topic.
\end{abstract}

    \begin{figure}
    \includegraphics[page=2,width=\textwidth]{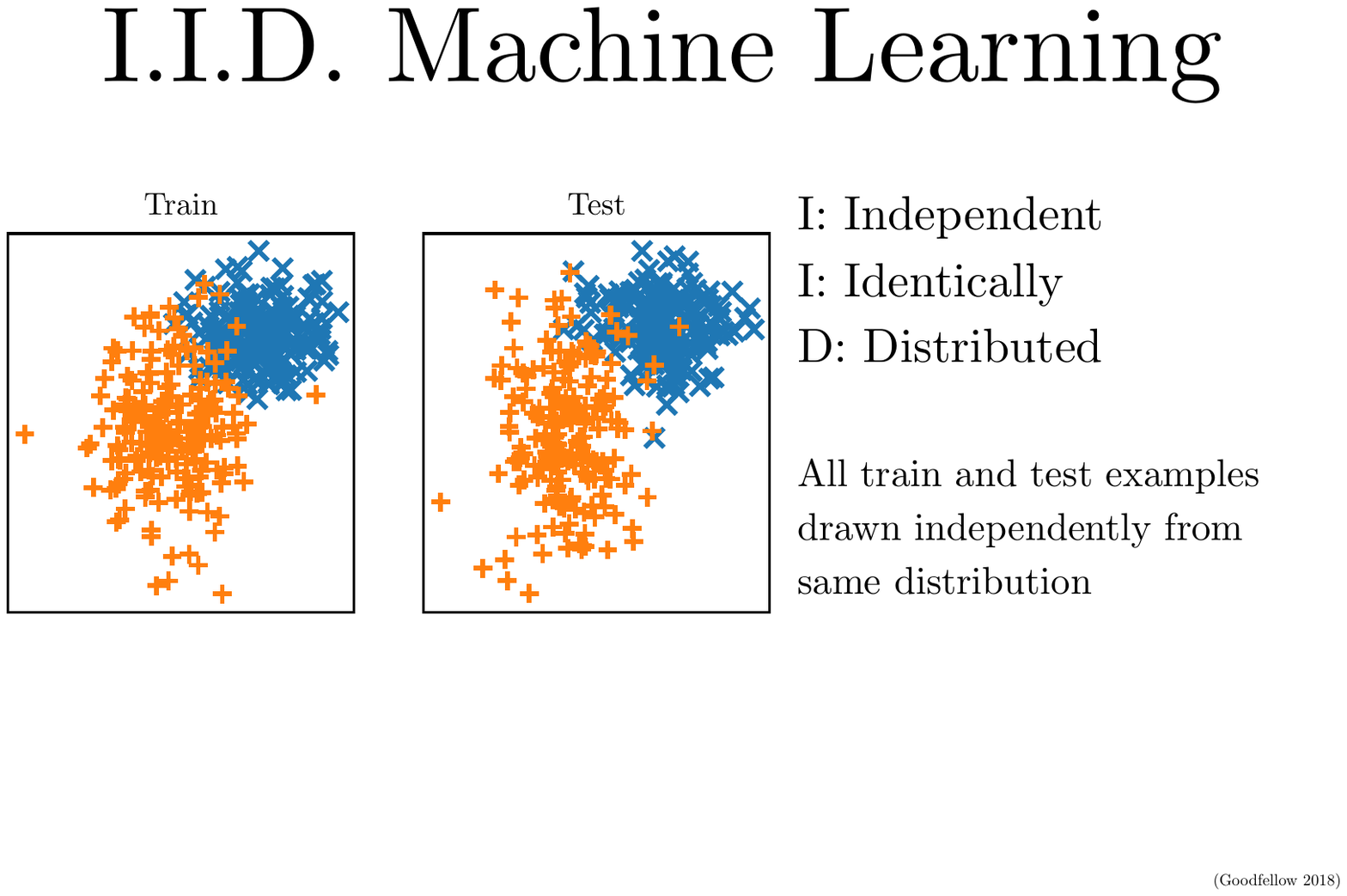}
            
            \caption[2]{Traditionally, most machine learning work has taken place in the context of the I.I.D. assumptions.

``I.I.D.'' stands for ``independent and identically distributed''. It means that all of the examples in the training and test set are generated independently from each other, and are all drawn from the same data-generating distribution.

This diagram illustrates this with an example training set and test set sampled for a classification problem with 2 input features (one plotted on horizontal axis, one plotted on vertical axis) and 2 classes (orange plus versus blue X).}
            
    \end{figure}
    
    \begin{figure}
    \includegraphics[page=3,width=\textwidth]{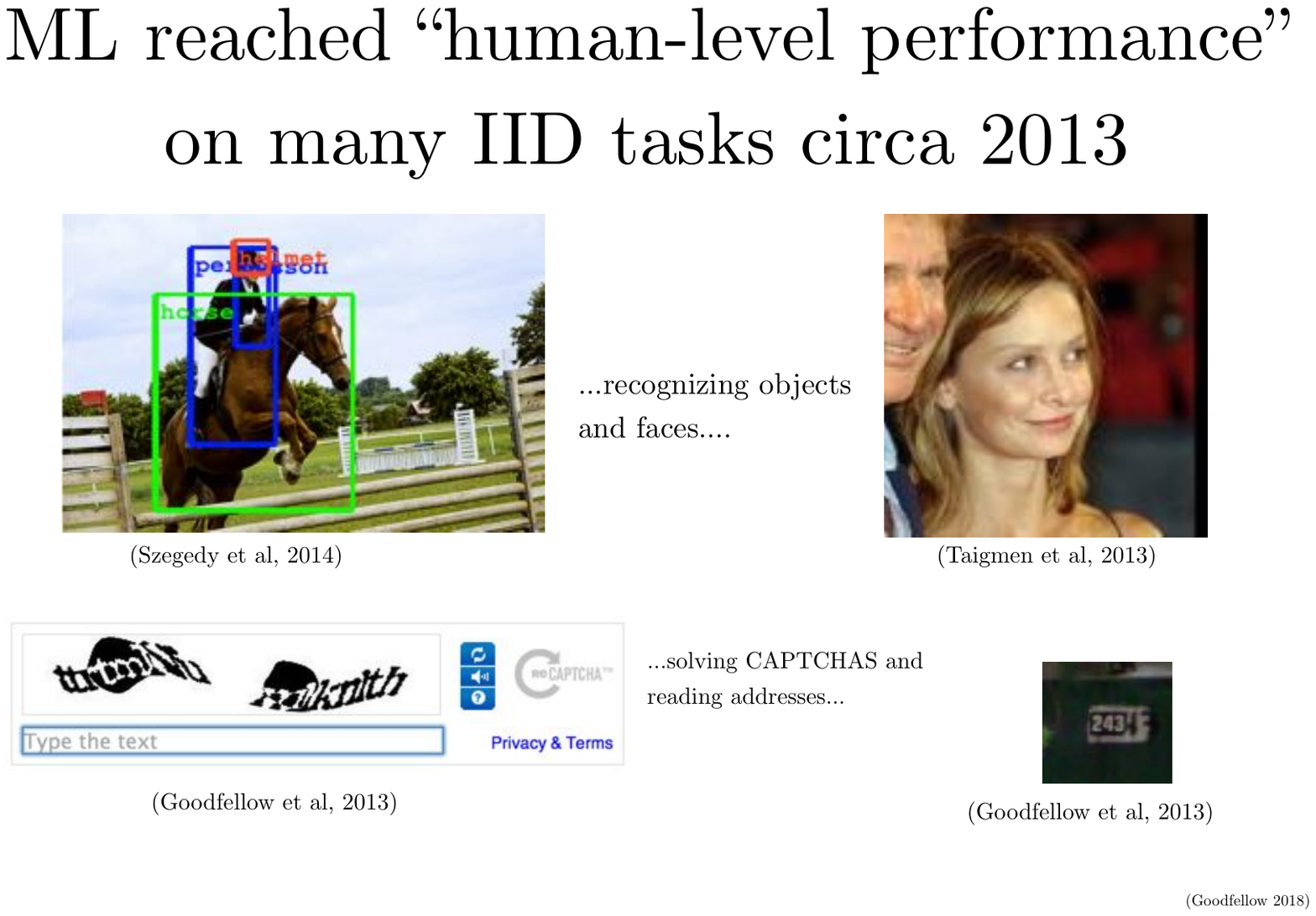}
            
            \caption[3]{Until recently, machine learning was difficult, even in the I.I.D. setting.

Adversarial examples were not interesting to most researchers because mistakes were the rule, not the exception.

In about 2013, machine learning started to reach human-level performance on several benchmark tasks (here I highlight vision tasks because they have nice pictures to put on a slide). These benchmarks are not particularly well-suited for comparing to humans, but they do show that machine learning has become quite advanced and impressive in the I.I.D. setting.}
            
    \end{figure}
    
    \begin{figure}
    \includegraphics[page=4,width=\textwidth]{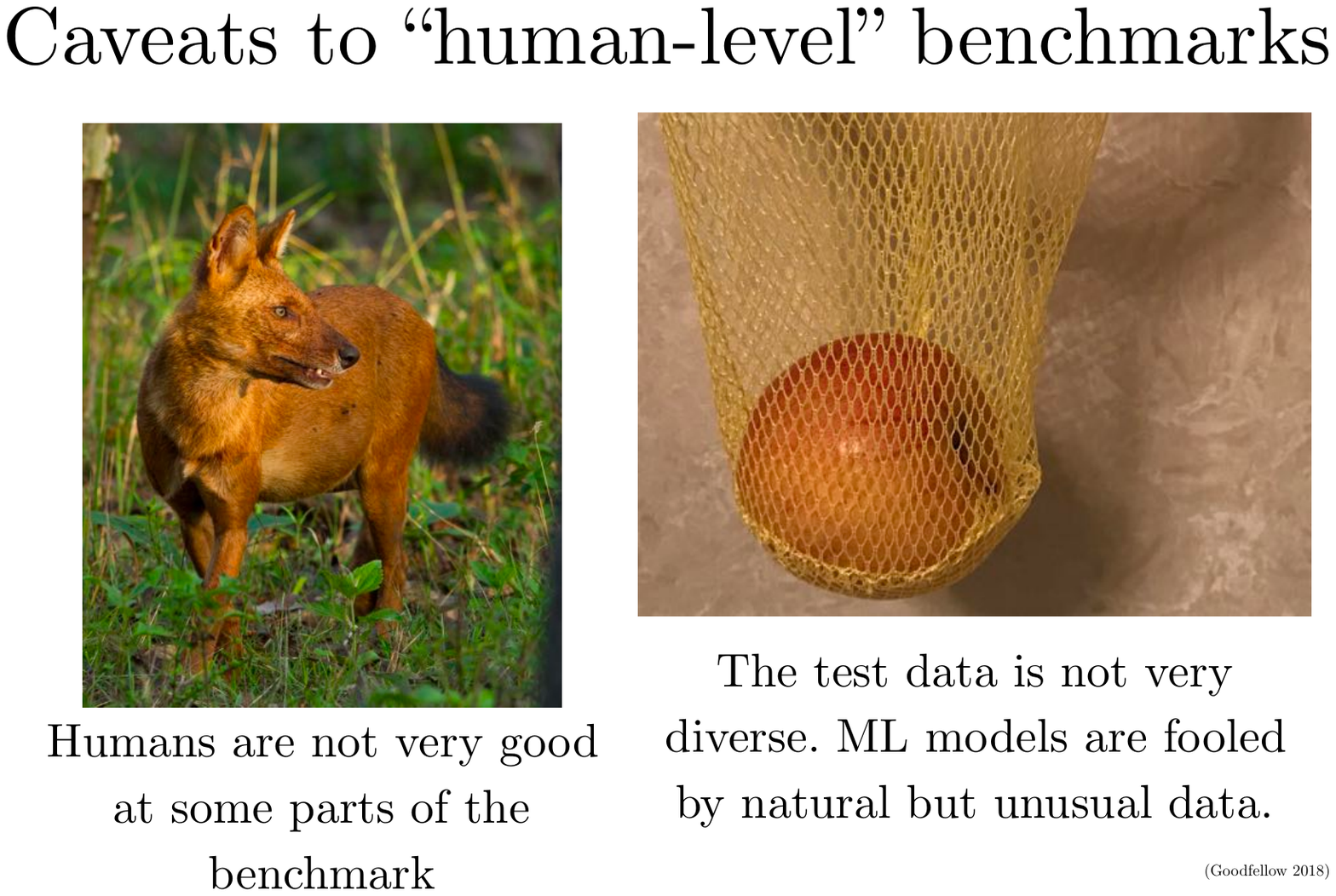}
            
            \caption[4]{When we say that a machine learning model has reached human-level performance for a benchmark, it is important to keep in mind that benchmarks may not be able to capture performance on these tasks realistically.

For example, humans are not necessarily very good at recognizing all of the obscure classes in ImageNet, such as this dhole (one of the 1000 ImageNet classes). Image from the Wikipedia article ``dhole''.

Just because the data is I.I.D. does not necessarily mean it captures the same
distribution the model will face when it is deployed.
For example, datasets tend to be somewhat curated, with relatively cleanly
presented canonical examples. Users taking photos with phones take unusual
pictures. Here is a picture I took with my phone of an apple in a mesh bag.
A state of the art vision model tags this with only one tag: ``material''.
My family wasn't sure it was an apple, but they could tell it was fruit and apple was their top guess. If the image is blurred the model successfully recognizes it as ``still life photography'' so the model is capable of processing this general kind of data; the bag is just too distracting.}
            
    \end{figure}
    
    \begin{figure}
    \includegraphics[page=5,width=\textwidth]{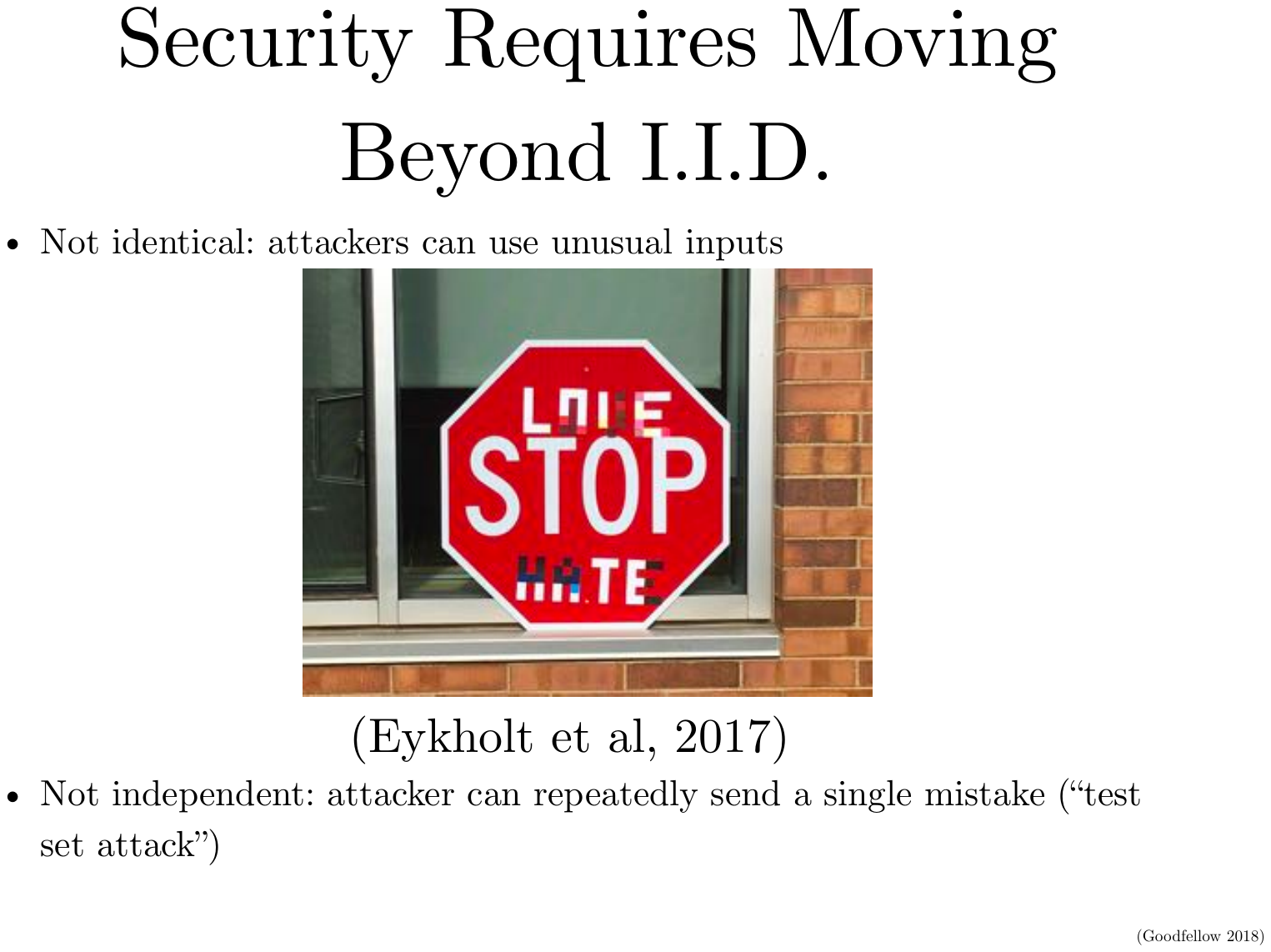}
            
            \caption[5]{When we want to provide security guarantees for a machine learning system, we can no longer rely on the I.I.D. assumptions.

In this presentation, I focus on attacks based on modifications of the input at test time. In this context, the two main relevant violations of the I.I.D. assumptions are:

The test data is not drawn from the same distribution as the training data. The attacker intentionally shifts the distribution at test time toward unusual inputs such as this adversarial stop sign ( https://arxiv.org/abs/1707.08945  ) that will be processed incorrectly.

The test examples are not necessarily drawn independently from each other. A real attacker can search for a single input that causes a mistake, and then send that input repeatedly, every time they interact with the system.
}
            
    \end{figure}
    
    \begin{figure}
    \includegraphics[page=6,width=\textwidth]{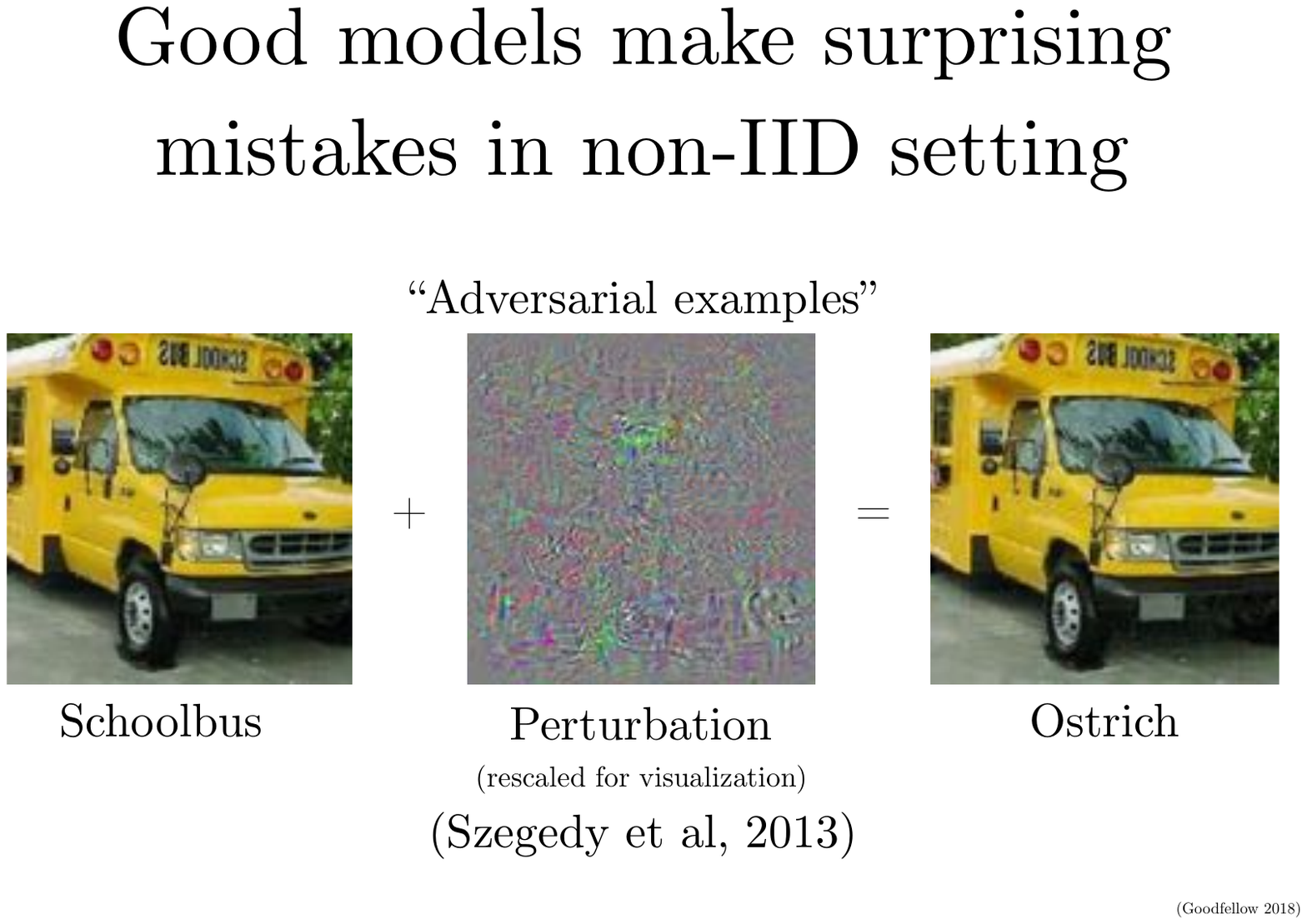}
            
            \caption[6]{The deep learning community first started to pay attention to surprising mistakes in the non-IID setting when Christian Szegedy showed that even imperceptible changes of IID test examples could result in consistent misclassification.

The paper ( https://arxiv.org/abs/1312.6199 ) introduced the term ``adversarial examples'' to describe these images. They were formed by using gradient-based optimization to perturb a naturally occurring image to maximize the probability of a specific class.

The discovery of these gradient-based attacks against neural networks was concurrent work  happening at roughly the same time as work done by Battista Biggio et al ( https://link.springer.com/chapter/10.1007\%2F978-3-642-40994-3\_25 ). Biggio et al's work was published earlier in 2013 while Christian's paper appeared on arxiv in late 2013. The first written record I personally have of Christian's work is a 2012 e-mail from Yoshua Bengio.}
            
    \end{figure}
    
    \begin{figure}
    \includegraphics[page=7,width=\textwidth]{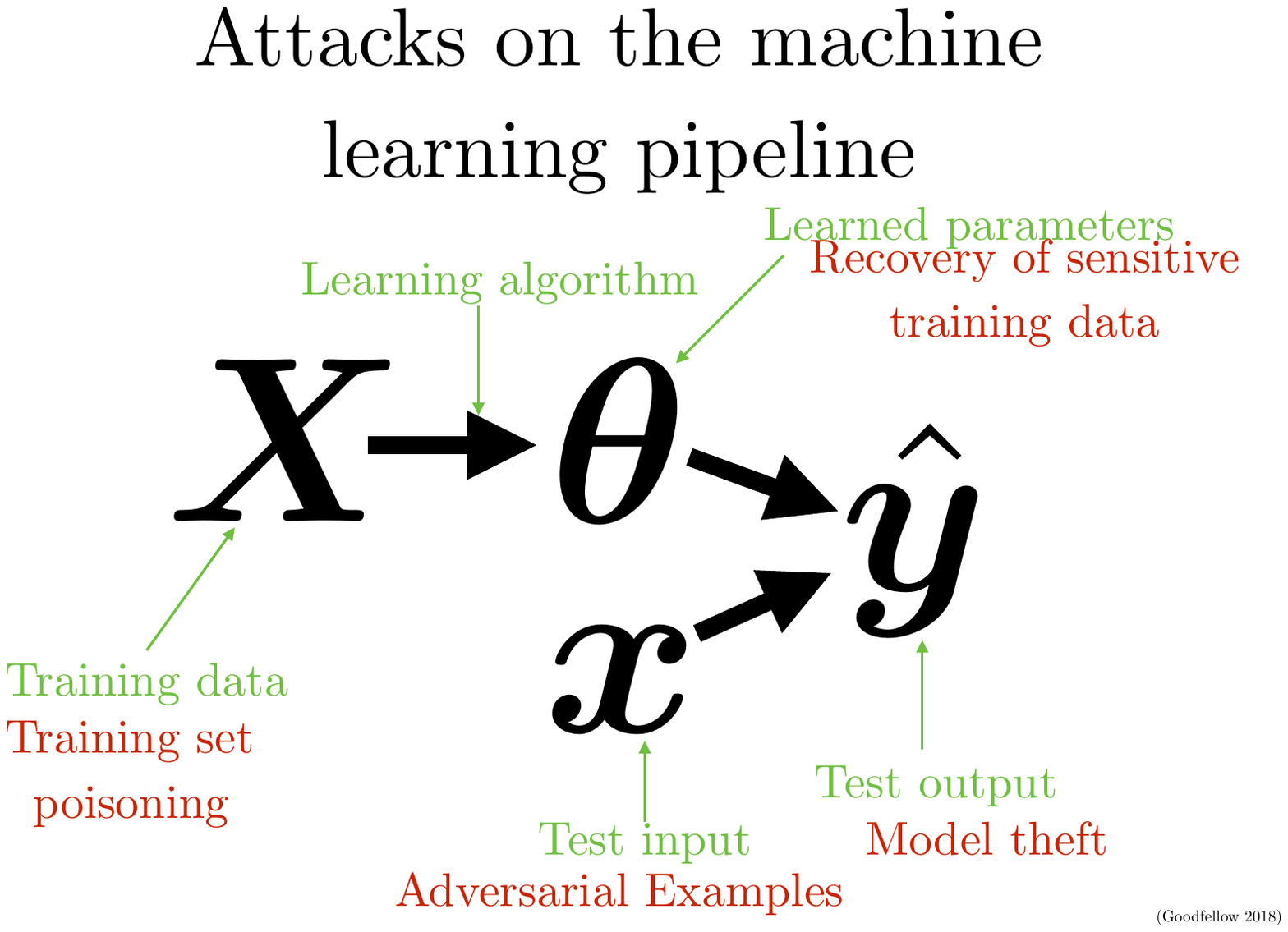}
            
            \caption[7]{To define adversarial examples more clearly, we should consider some other security problems.
Many machine learning algorithms can be described as a pipeline that takes training data X, learns parameters theta, and then uses those parameters to process test inputs x to produce test outputs y-hat.

Attacks based on modifying the training data to cause the model to learn incorrect behaviors are called training set poisoning.

Attackers can study learned parameters theta to recover sensitive information from the training set (for example, recovering social security numbers from a trained language model as demonstrated by https://arxiv.org/abs/1802.08232 ).
Attackers can send test inputs x and observe outputs y-hat to reverse engineer their model and train their own copy. This is known as a model theft attack. Model theft can then enable further attacks, like recovery of private training data or improved adversarial examples.

Adversarial examples are distinct from these other security concerns: they are inputs supplied at test time, intended to cause the model to make a mistake.}
            
    \end{figure}
    
    \begin{figure}
    \includegraphics[page=8,width=\textwidth]{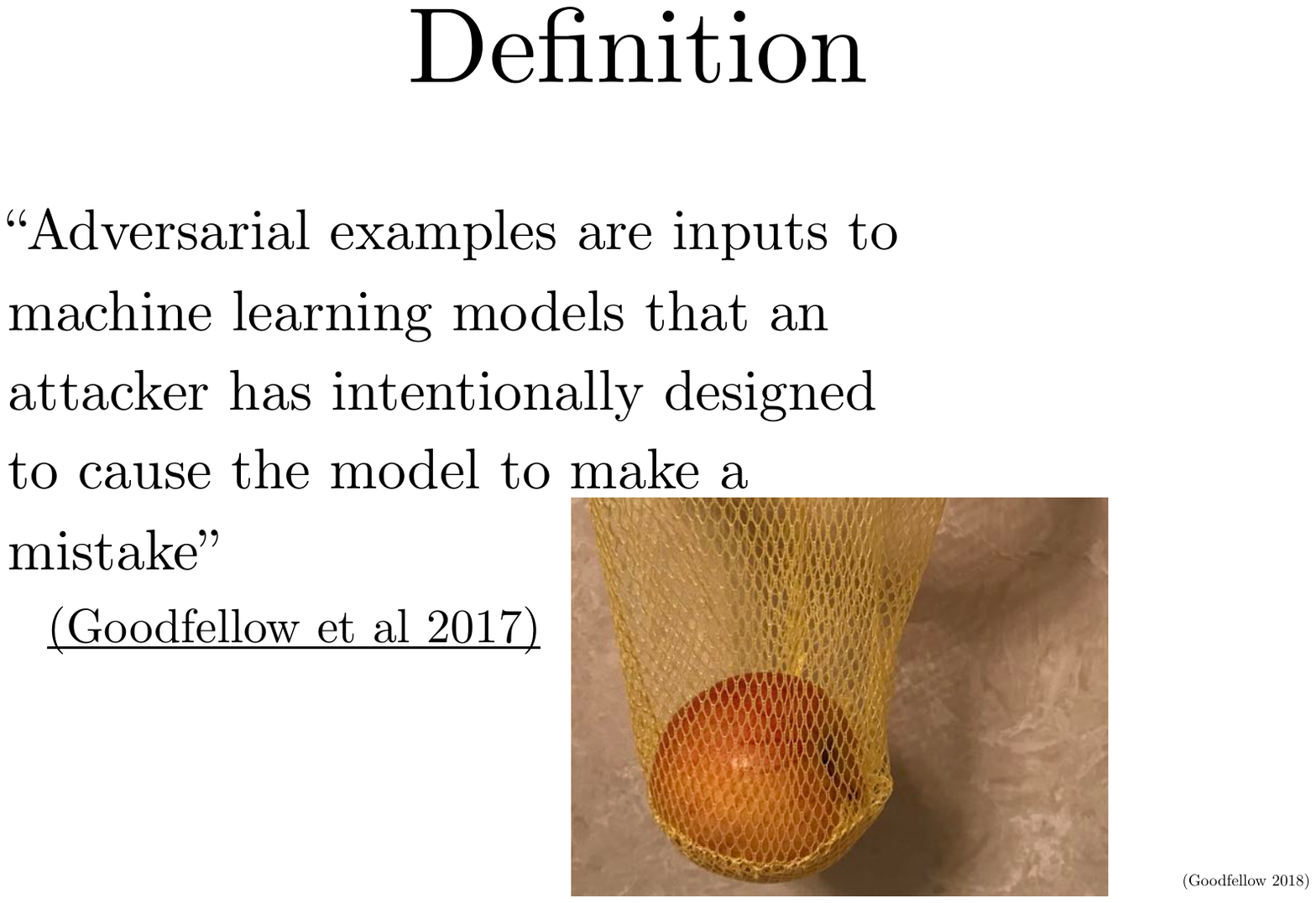}
            
            \caption[8]{There is no standard community-accepted definition of the term ``adversarial examples'' and the usage has evolved over time.

I personally coined the term ``adversarial examples'' while helping to write Christian's paper so I feel somewhat within my rights to push a definition.

The definition that I prefer today was introduced in an OpenAI blog post and developed with my co-authors of that blog post.

There are three aspects of this definition I want to emphasize.

There is no need for the adversarial example to be made by applying a small or imperceptible perturbation to a clean image. That was how we used the term in the original paper, but its usage has evolved over time. In particular, the picture of the apple in the mesh bag counts. I went out of my way to find a strange context that would fool the model.

2) Adversarial examples are not defined in terms of deviation from human perception, but in terms of deviation from some absolute standard of correct behavior. In contexts like visual object recognition, human labelers might be the best approximation we have to the ground truth, but human perception is not the definition of truth. Humans are subject to mistakes and optical illusions too, and ideally we could make a machine learning system that is harder to fool than a human.

3) An adversarial example is intended to be misclassified, but the attacker does not necessarily succeed. This makes it possible to discuss ``error rate on adversarial examples''. If adversarial examples were defined to be actually misclassified, this error rate would always be 1 by definition.

For a longer discussion see https://arxiv.org/abs/1802.08195}
            
    \end{figure}
    
    \begin{figure}
    \includegraphics[page=9,width=\textwidth]{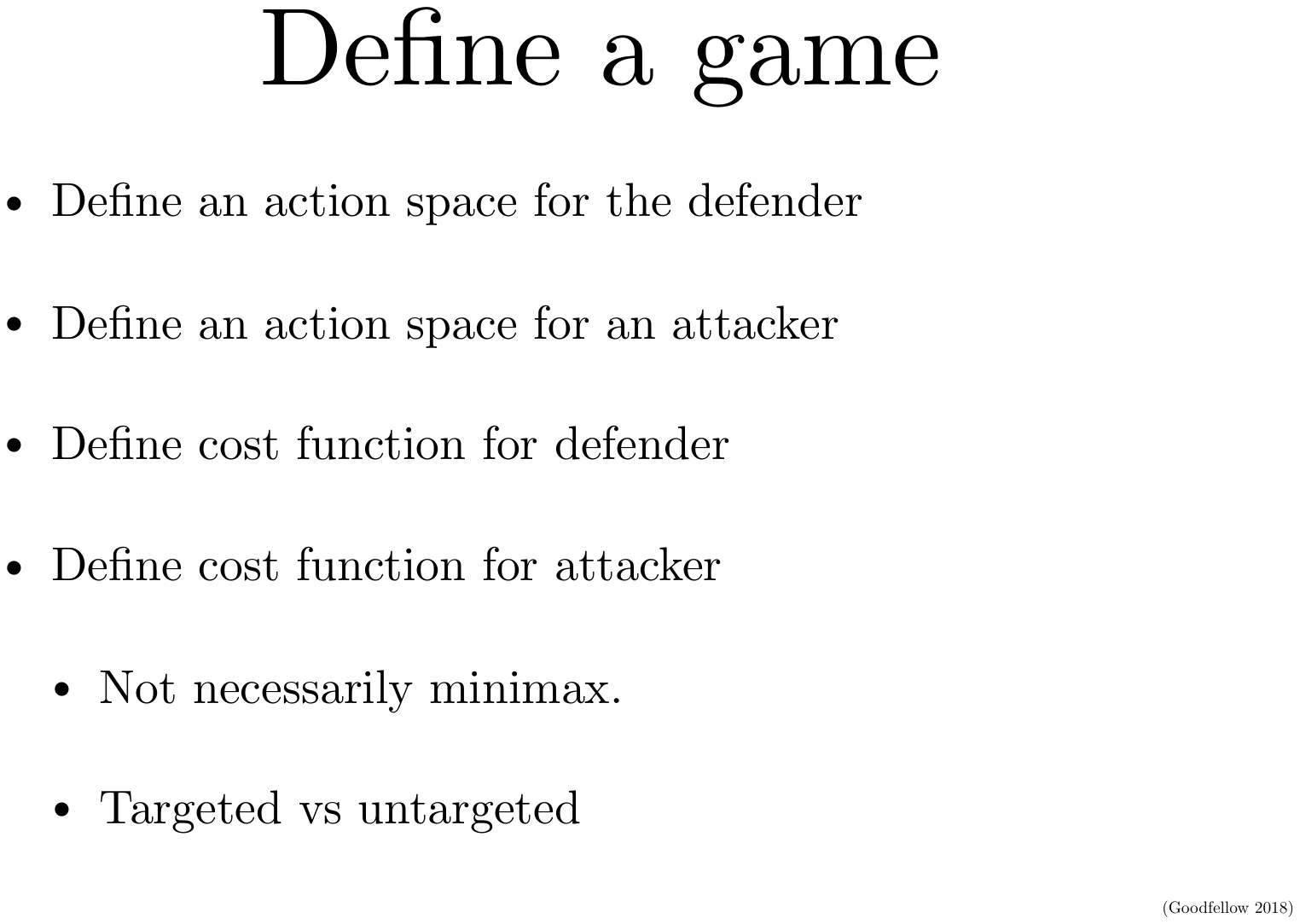}
            
            \caption[9]{To study machine learning in the adversarial setting, we must define a game, formally.

This means we must define an action space and cost function for both the attacker and the defender.

Usually, the defender's action space is to output a class ID, but we can also imagine other variants of the game, where the defender can output a confidence value or can refuse to classify adversarially manipulated inputs.

In the context of adversarial examples, the attacker's action space describes the kind of inputs that the attacker can present to the defender's model.

The defender's cost function, for the purpose of the game, is usually some kind of error rate. Note that this is different from the cost used to train the neural net, which is designed with other concerns like differentiability in mind. The cost for the purpose of the game should directly measure the actual performance of the defender.

Many people often think of adversarial settings as necessarily involving minimax games, but that is not always the case. In a minimax game, the attacker's cost is just the negative cost of the defender. Other cost functions for the attacker often make sense. For example, the defender may want to get as many examples correct as possible while the attacker may gain an advantage only from causing specific mistakes. An untargeted attacker just wants to cause mistakes, but a targeted attacker wants to cause an input to be recognized as coming from a specific class. For example, to sneak into a secure facility by fooling face recognition, it is not enough for the attacker to fool the face recognition system into guessing their identity incorrectly. The attacker must be recognized specifically as an individual who has access to the facility.

}
            
    \end{figure}
    
    \begin{figure}
    \includegraphics[page=10,width=\textwidth]{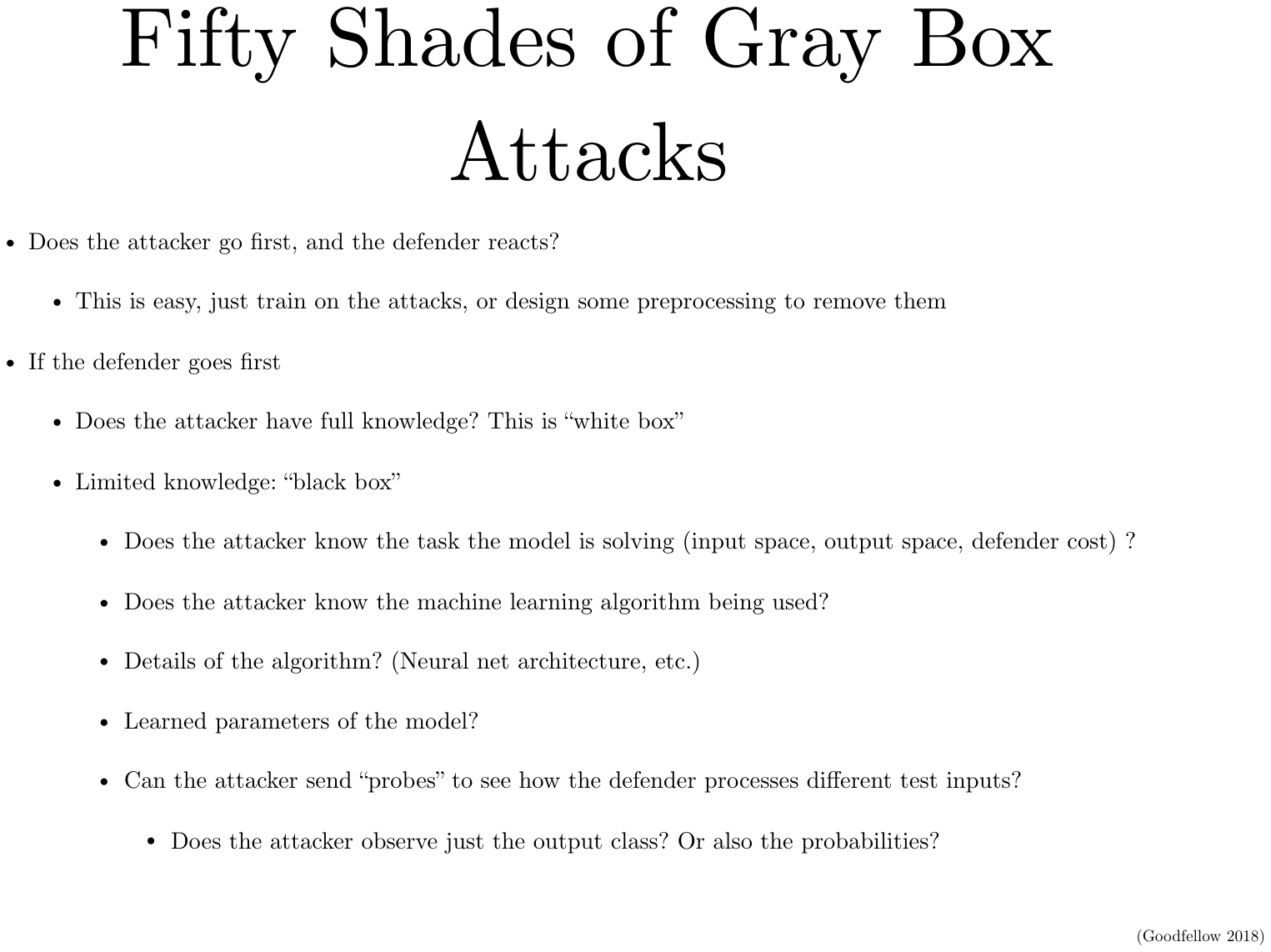}
            
            \caption[10]{
Another important part of specifying the game is specifying the amount
of access that the attacker has to the model and whether the attacker
or the defender makes the first move in the game.
If we believe that the defender first specifies a machine learning
algorithm and then the attacker specifies an attack algorithm,
the defender must provide a defense that can adapt to new attack
algorithms.
If the attacker has full access to the model, this is called a
``white box'' scenario, while if the attacker has no information
about the model, this is called a ``black box scenario''.
Note that the white box scenario implies that the defender moves first.
Many papers do not explicitly describe which player moves first,
but by studying the white box scenario they implicitly study games
where the defender moves first.
The term ``black box'' is used inconsistently in the literature;
papers that study ``black box'' scenarios should explicitly specify
the amount of access they allow, since many of these papers allowing
differing amounts of ``gray box'' access.
}
            
    \end{figure}
    
    \begin{figure}
    \includegraphics[page=11,width=\textwidth]{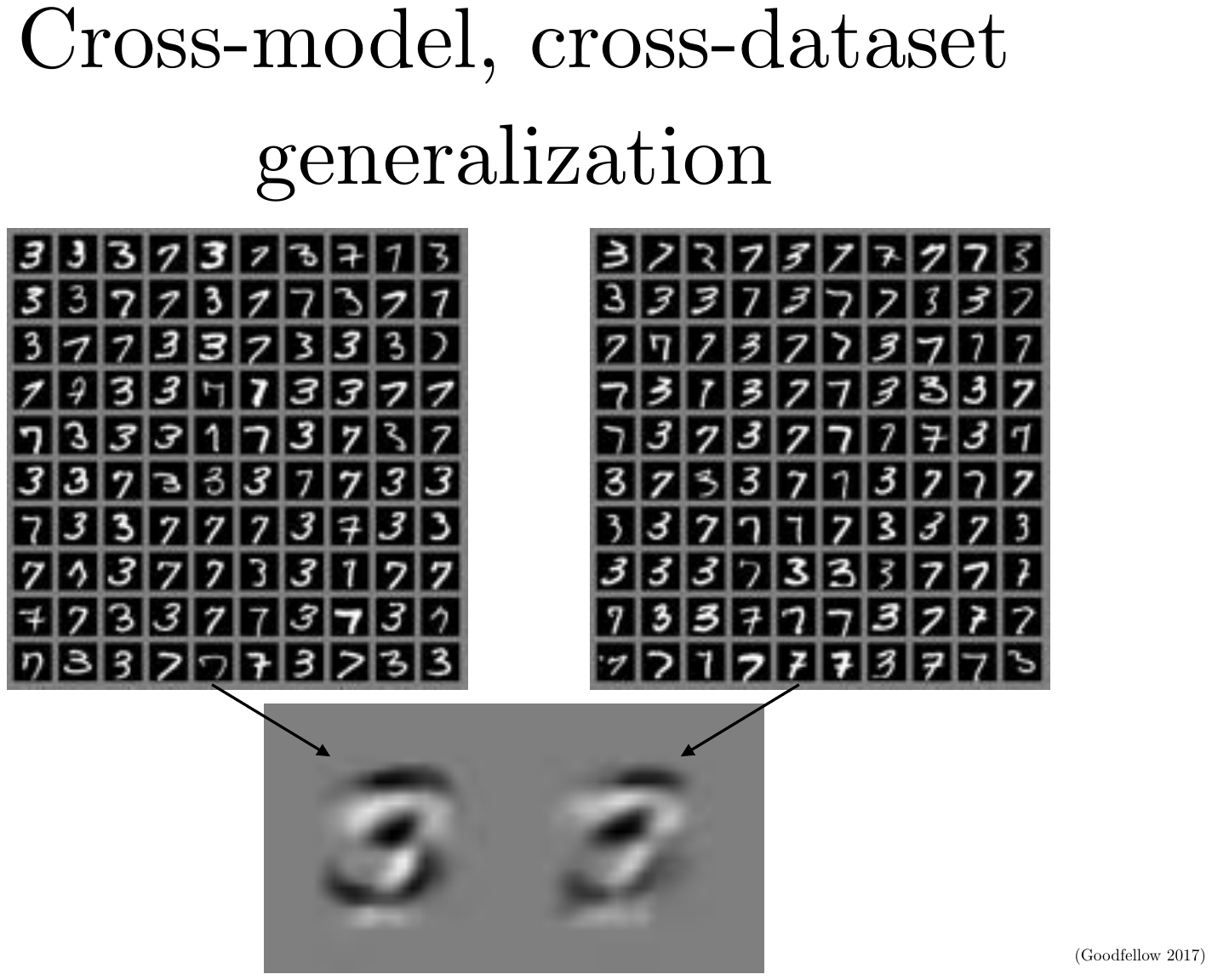}
            
            \caption[11]{Of particular concern for security purposes
is the fact that adversarial examples transfer from one model
to another, first observed by Szegedy et al 2013.
In other words, adversarial examples that fool one model are
likely to fool another. This is the basis for many black box
attack strategies.
This figure shows how two logistic regression models trained
on two different subsets of the MNIST 3s and 7s learn similar
weights and thus will have similar adversarial examples.
}
            
    \end{figure}
    
    \begin{figure}
    \includegraphics[page=12,width=\textwidth]{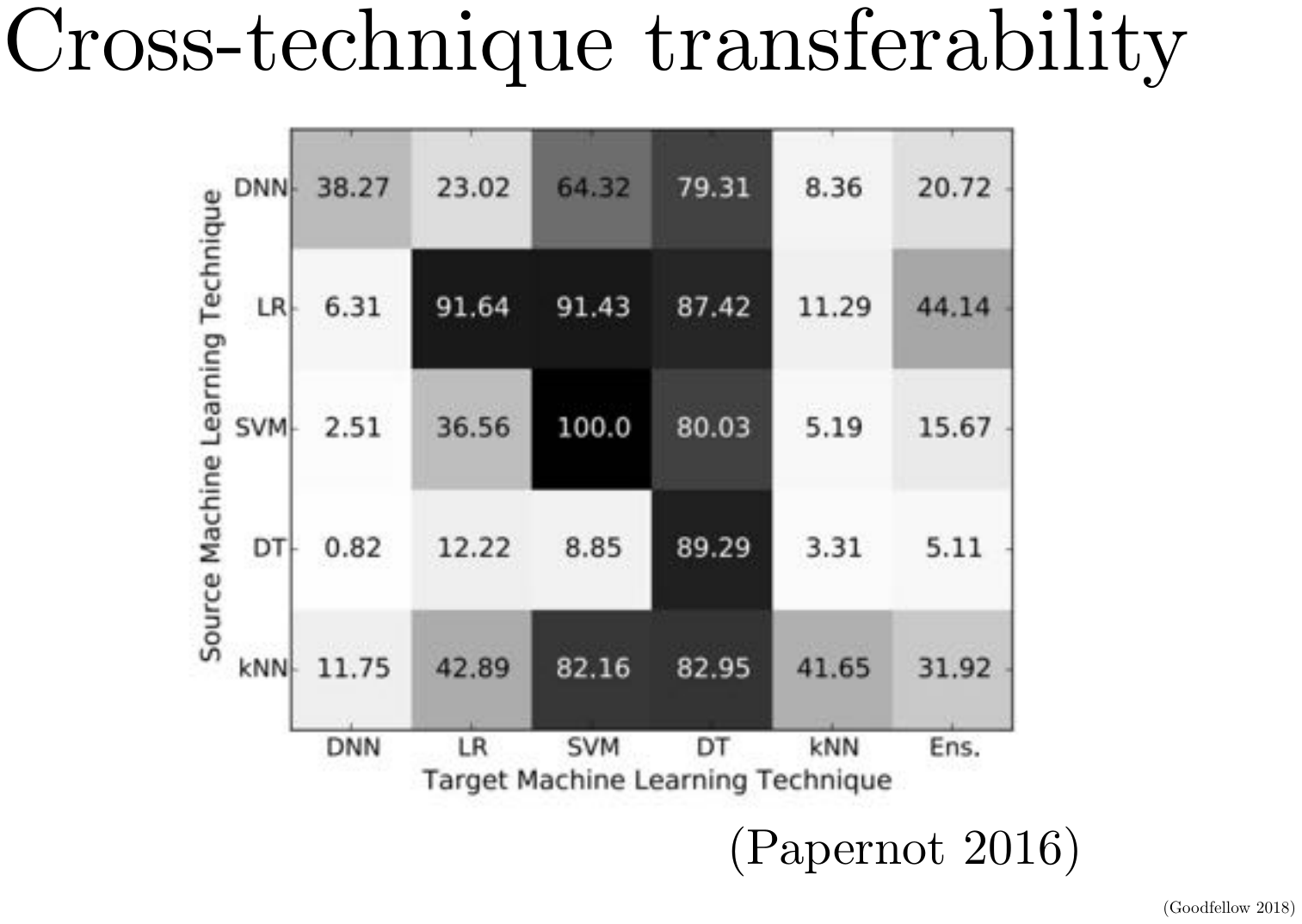}
            
            \caption[12]{Adversarial examples often transfer between
entirely different kinds of machine learning algorithms, such
as SVMs and Decision Trees.}
            
    \end{figure}
    
    \begin{figure}
    \includegraphics[page=13,width=\textwidth]{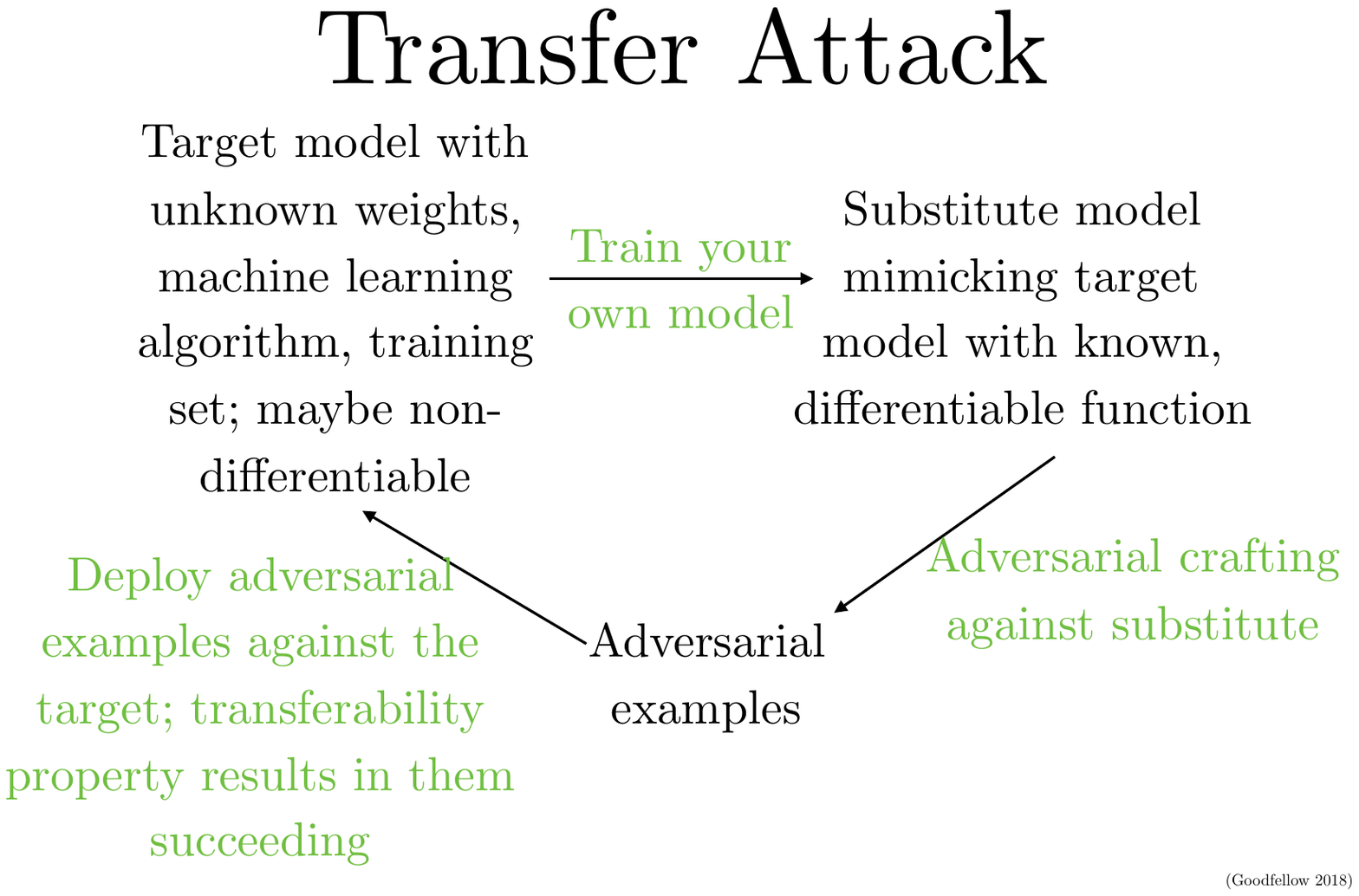}
            
            \caption[13]{A major class of black-box attacks is based
on exploiting transfer between models.
The attacker constructs their own model, either by training
their own model on their own training set, or (in threat models
that allow this) by sending inputs
to the target model and training their own model to mimic its
outputs. Adversarial examples for the attackers model then transfer
to the target model.

When the attacker trains their own model on their own data, the
attacker relies somewhat on luck for the adversarial examples to
transfer to the target, because the transfer rate is not perfect.
When the attacker trains their model to mimic the target model,
as done by Papernot et al 2016, the transfer rate improves as
the substitute model learns the decision boundary of the target
model more accurately, so the attacker is able to actually invest
effort and resources to boost the success rate of the attack.}
            
    \end{figure}
    
    \begin{figure}
    \includegraphics[page=14,width=\textwidth]{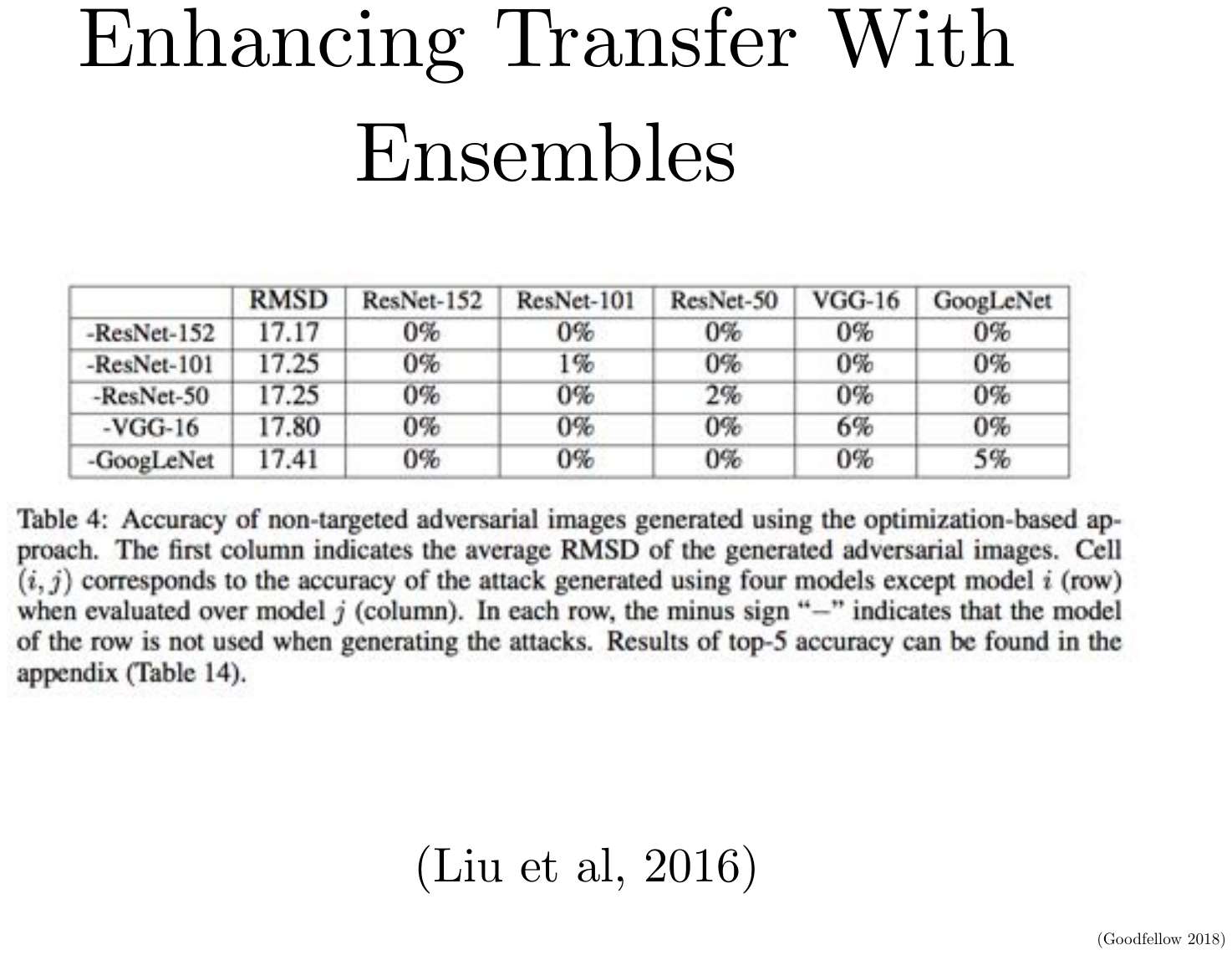}
            
            \caption[14]{Attackers operating in the purely black box setting
who cannot send inputs to the target model in order to reverse-engineer
it can boost their chances of success by finding adversarial examples
that fool an entire ensemble of models trained by the attacker.
In this experiment, adversarial examples that fooled five substitute
models had a 100\% chance of fooling the target model.}
            
    \end{figure}
    
    \begin{figure}
    \includegraphics[page=15,width=\textwidth]{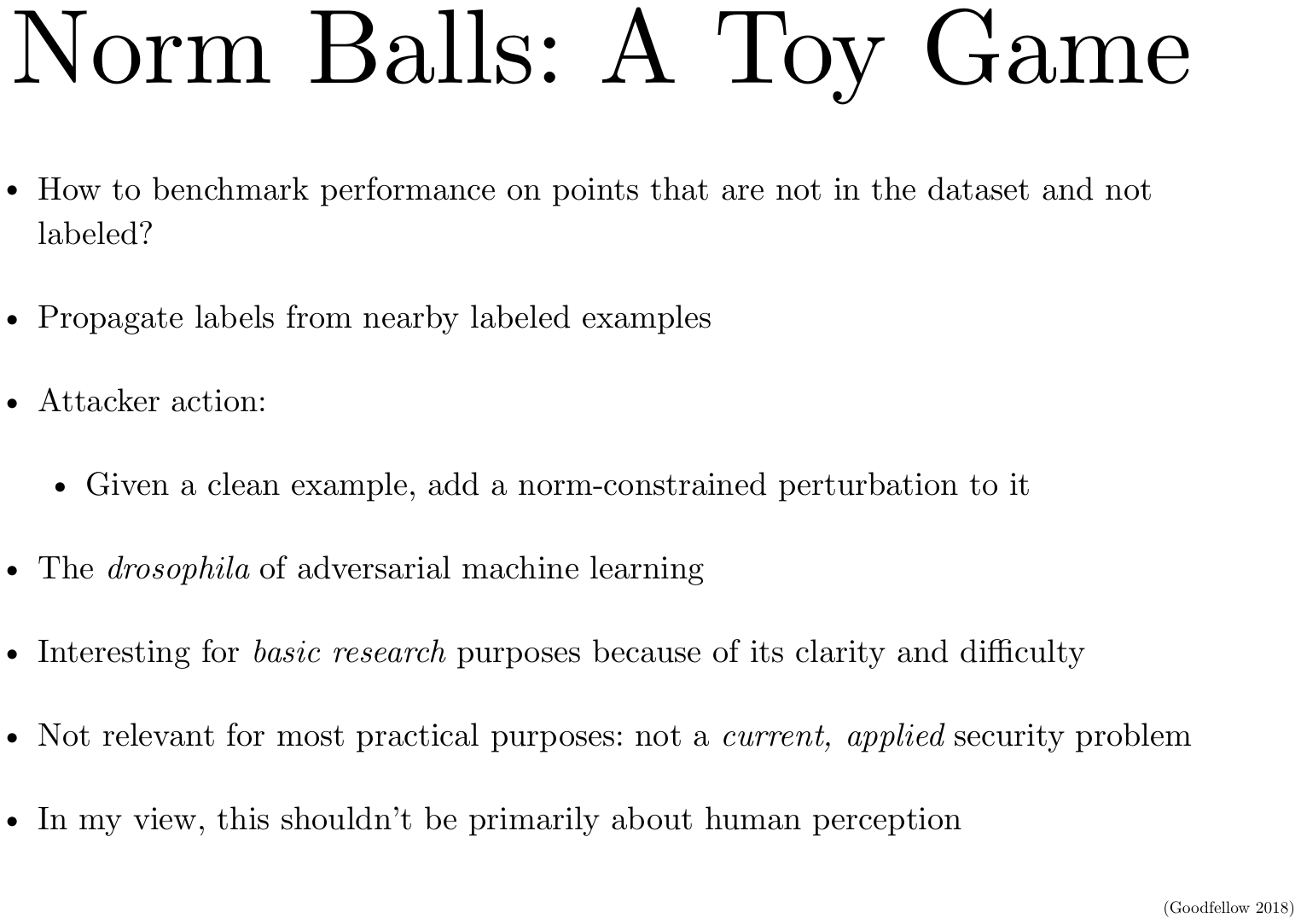}
            
            \caption[15]{Most adversarial example research today is based on a specific toy game in the context of visual object recognition.

We want to evaluate the performance of a classifier on arbitrary inputs, since in most scenarios the attacker is not constrained to supply naturally occurring data is input. Unfortunately, for visual object recognition, it is not straightforward to evaluate the classifier on arbitrary inputs. We rely on human labelers to obtain ground truth labels, and it is slow and expensive to include a human in the loop to label all attack images.

To obtain an inexpensive and automated evaluation, we can propagate labels from nearby points ( https://arxiv.org/abs/1412.6572 ). This suggests that for research purposes, a convenient game to study is one where the attacker's action space is to take a clean test image and modify it by adding a norm-constrained perturbation. The size epsilon of this perturbation is chosen to ensure that the resulting adversarial examples have the same class as the clean examples. Epsilon should be made as large as possible while still preserving classes in order to benchmark performance on as large a subset of the input space as possible.

These games have gained a lot of attention because, despite their simplicity, it has been extremely difficult for a defender to win such a game.

However, it is important to emphasize that these games are primarily tools for basic research, and not models of real-world security scenarios. One of the best things that one could do for adversarial machine learning research is to devise a practical means of benchmarking performance in more realistic scenarios.}
            
    \end{figure}
    
    \begin{figure}
    \includegraphics[page=16,width=\textwidth]{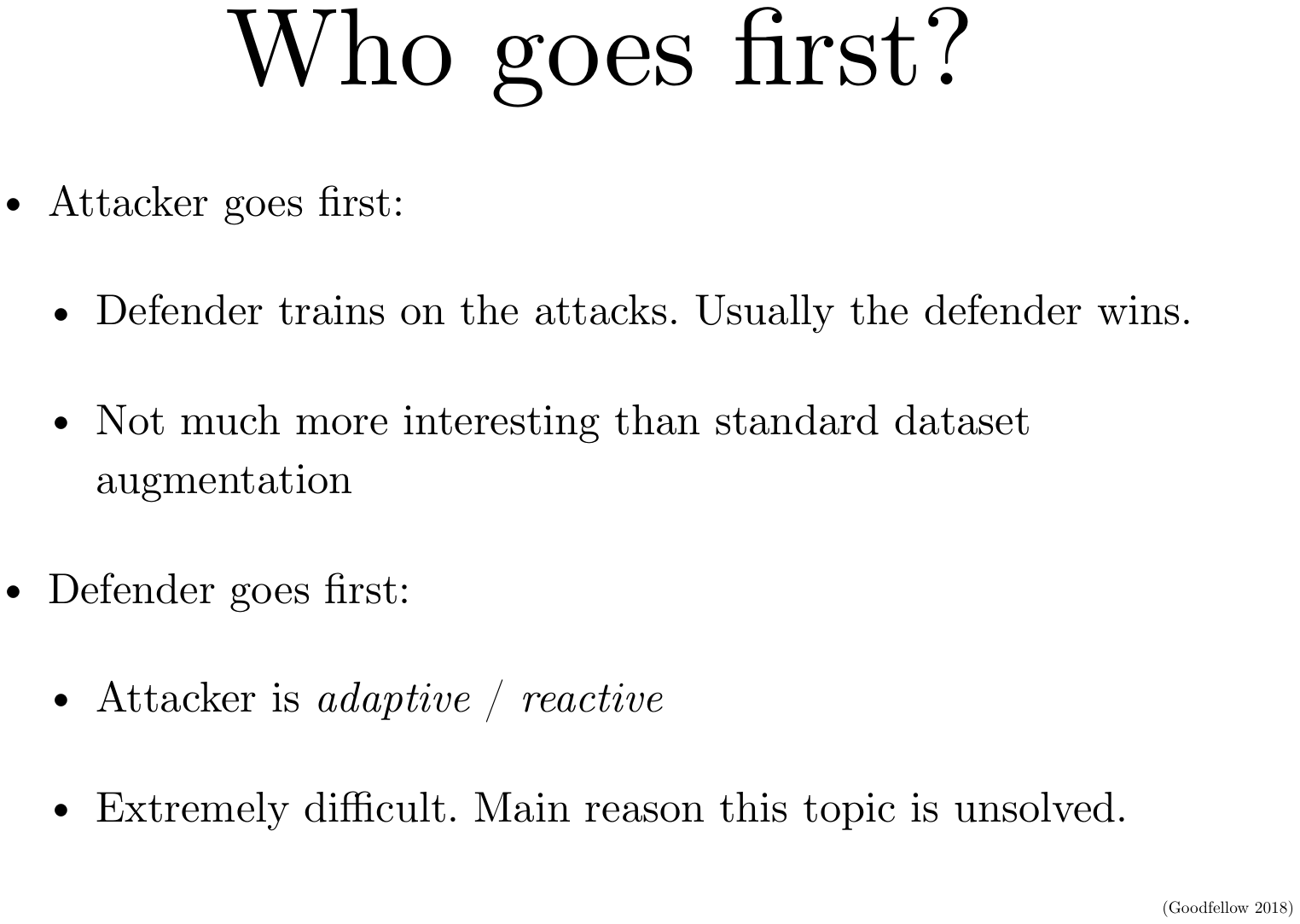}
            
            \caption[16]{An example of where the person trying to fool the neural net goes first is text CAPTCHAs. Text-based CAPTCHAs have been broken since 2013. https://static.googleusercontent.com/media/research.google.com/en//pubs/archive/42241.pdf}
            
    \end{figure}
    
    \begin{figure}
    \includegraphics[page=17,width=\textwidth]{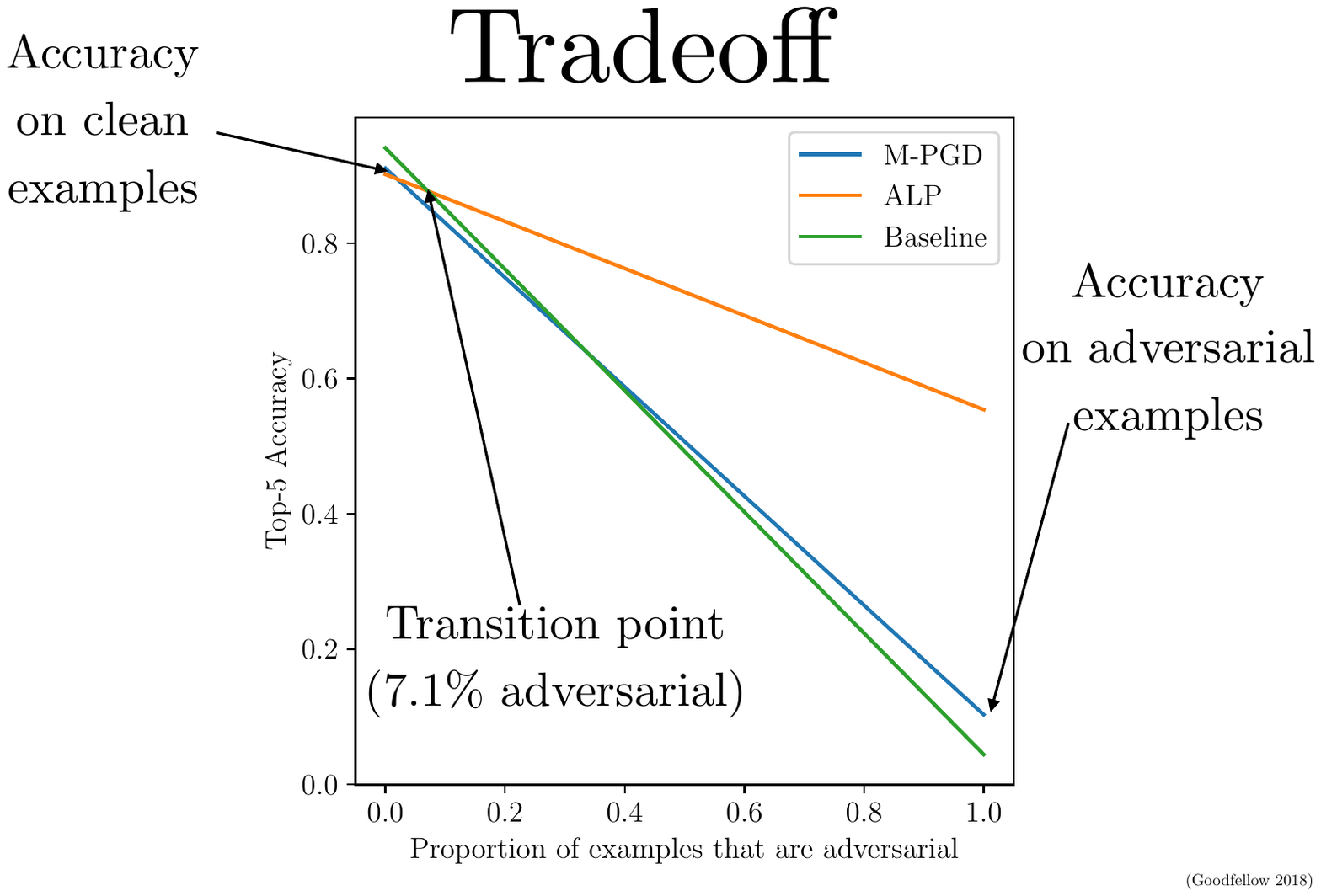}
            
            \caption[17]{In some cases, defenses against adversarial examples act as a regularizer and actually improve accuracy on the test set ( https://arxiv.org/abs/1412.6572 ). p p
In most of the recent literature, the strongest defenses against adversarial examples tend to decrease accuracy on the clean test set.
To choose a specific model to use on a particular task, we shoul p pd consider the cost it will incur when it is actually used in practice. To simplify the discussion, assume all errors are equally costly. If we believe a model will always receive adversarial examples, we should choose the model with the highest accuracy in the adversarial setting. If we believe it will receive only clean examples, then we should choose the model with the highest accuracy on the clean test set. In many settings, we probably expect the model to come under attack a certain percentage of the time. If we assume that this percentage is independent of the choice of model (in reality, models with greater vulnerability may be attacked more often) we should choose the model that performs the best on a test set consisting of this proportion of adversarial examples.
For example, this plot shows the accuracy of an undefended baseline and two defenses, M-PGD and ALP ( https://arxiv.org/abs/1803.06373 ) on the ImageNet test set. The ALP model is more robust to adversarial examples, but at the cost of accuracy on the clean test set. For the ALP model to be preferable to the baseline, we need to expect that the model will face adversarial examples about 7.1\% of the time on the test set. Another interesting thing we see from this plot is that it shows us some tradeoffs are just not worth it: M-PGD might look like it offers complementary advantages and disadvantages relative to ALP because it has higher accuracy on the clean test set, but there is actually never any point on the curve where M-PGD has the highest accuracy. We can optimally navigate the tradeoff with only ALP and the undefended baseline as our two choices.
The choice between these two models is complicated by the fact that accuracies in the adversarial setting are usually upper bounds, based on testing the defense against a particular attack algorithm. A smarter attack algorithm or an attack using a different threat model could bring the accuracy of either model even lower in the adversarial setting, so there is uncertainty about the true tradeoff.
Of course, for the purposes of basic research, it still makes sense to study how to obtain better accuracy in the completely adversarial setting, but we must not lose sight of the need to retain good performance on clean data.}
            
    \end{figure}
    
    \begin{figure}
    \includegraphics[page=18,width=\textwidth]{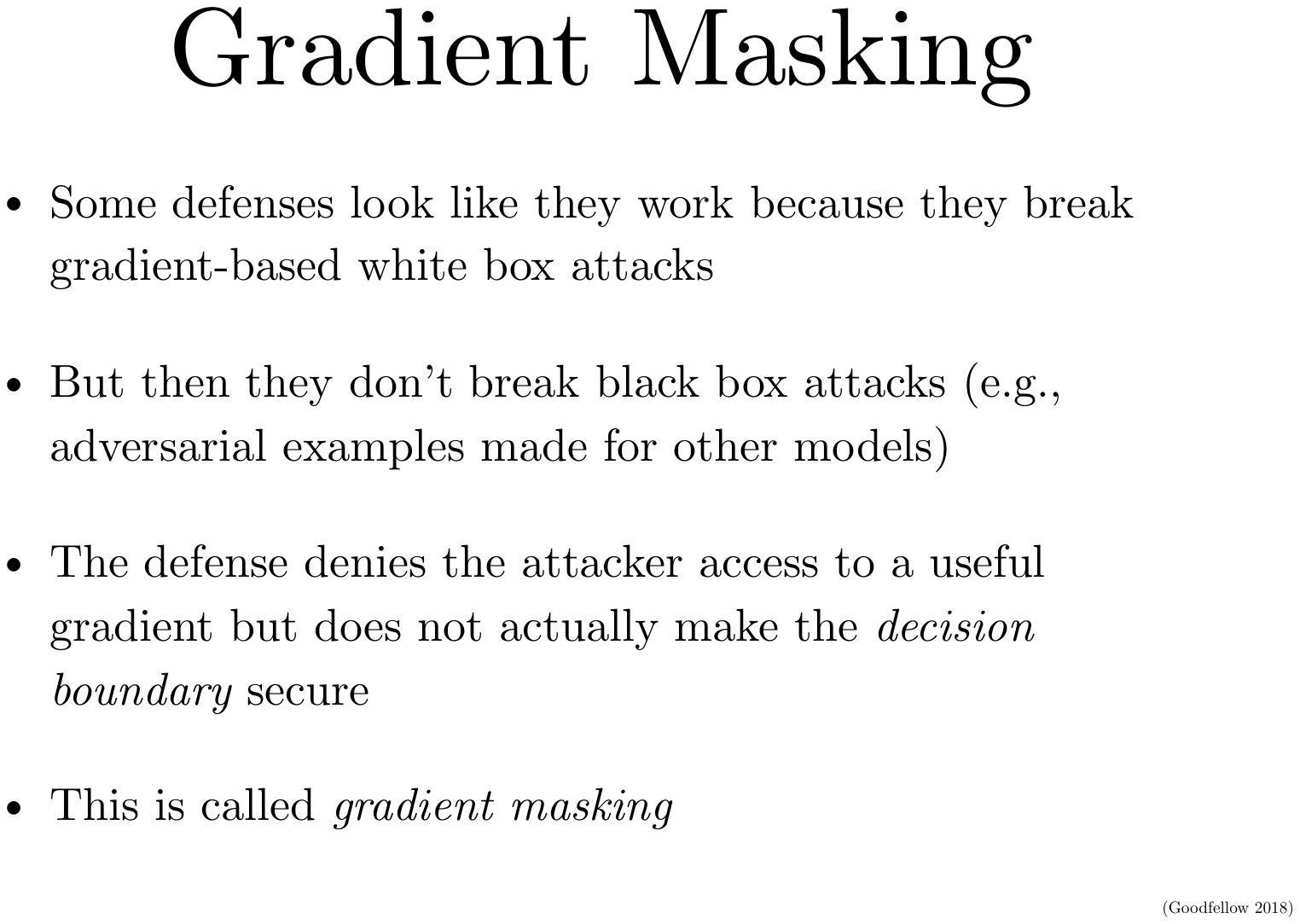}
            
            \caption[18]{For further reading:
https://arxiv.org/abs/1602.02697
https://arxiv.org/abs/1705.07204
https://arxiv.org/abs/1802.00420}
            
    \end{figure}
    
    \begin{figure}
    \includegraphics[page=19,width=\textwidth]{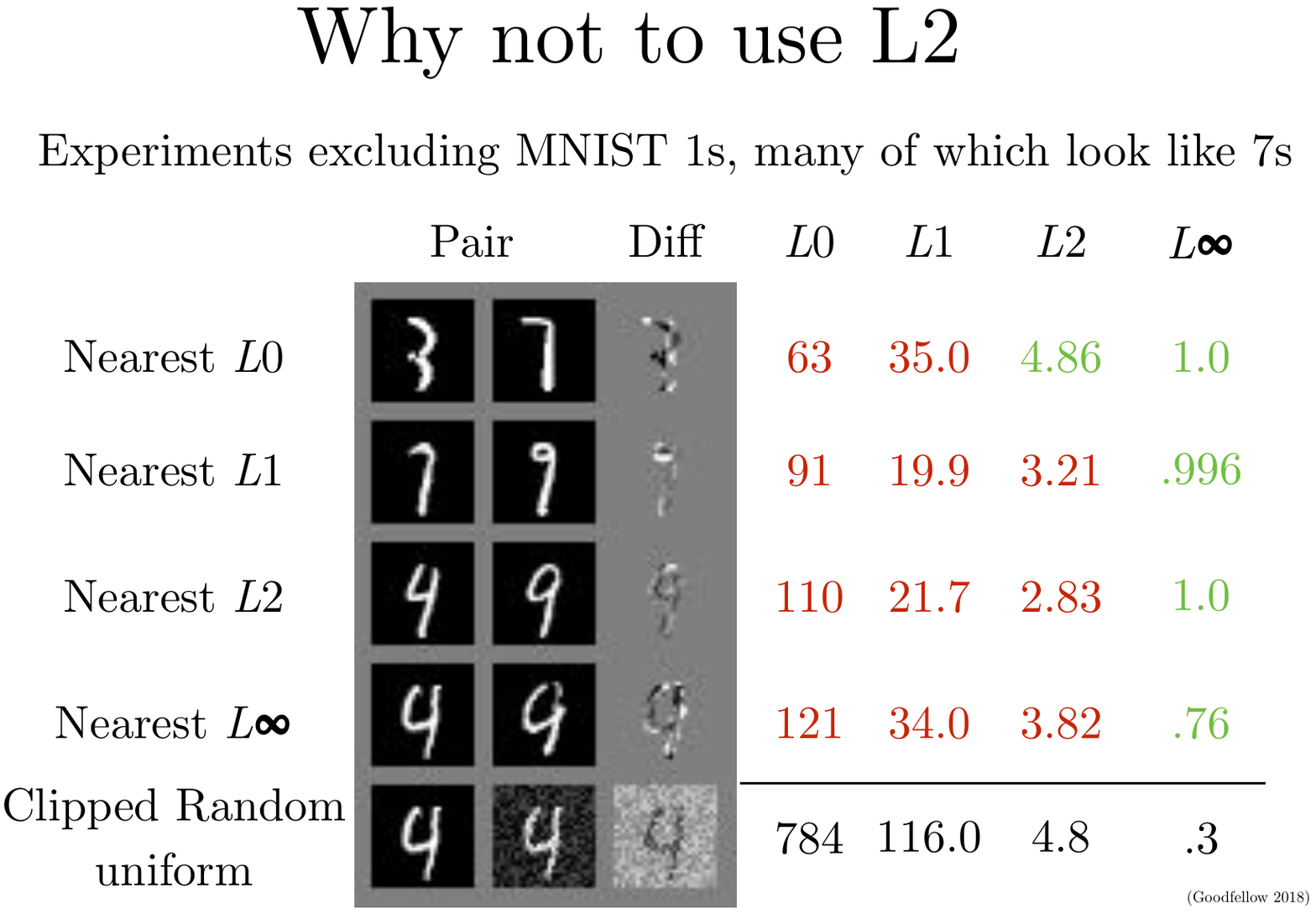}
            
            \caption[19]{To propagate labels from points in the dataset with known labels to nearby off-dataset points with unknown labels, we need some way to measure distance. In most current work on adversarial examples, this is done with the $L^\infty$ norm, advocated by https://arxiv.org/abs/1412.6572

This is intended to be a way of guaranteeing that the label is known on new test points.
Ideally we would like to propagate labels to as large a volume of space as possible.
(A common misconception is that we want to keep the perturbations small, to be
imperceptible---actually we would like to benchmark on all of $\mathbb{R}^n$ if we had a way of labeling it)

Norms are convenient to implement and to study mathematically, but some norms are better than others for propagating labels. This is of course highly application-specific. The $L^\infty$ norm is relevant primarily for visual object recognition tasks. For other tasks like malware detection, we would be interested in transformations of code that preserve its function.

In this example, we see that if we want to add large uniform noise (within the confines of the unit hypercube), the $L^\infty$ norm is the best at assigning larger distances to noisy perturbations than to perturbations that change the class. L0, L1, and L2 all assign smaller distances to examples that lie in different classes than to noisy versions of the example shown. The $L^\infty$ does not do this. We also see that if we constraint the input using the $L^\infty$ norm, we can get relatively large perturbations in terms of the other norms. Our $L^\infty$-constrained uniform perturbation has an L2 norm larger than most of the class-changing perturbations shown here. Intuitively, restricting the perturbation using the $L^\infty$ makes sure that the adversary cannot focus the whole perturbation on a small number of pixels, to completely erase or completely draw in ink that changes the MNIST digit.

The example of uniform noise makes L0, L1, and L2 all look bad, but L0 and L1 can perform better in other examples. It is mostly L2 that I intend to discourage here.

It would be great if researchers could find an improved method of reliably propagating labels to more points in space than this norm-ball approach allows. It is important to remember that the goal to an improved evaluation should either be to label more points or to more realistically model an actual security threat. In particular, the goal is not to find a good model of human perceptual distance, unless that helps with either of the preceding goals.}
            
    \end{figure}
    
    \begin{figure}
    \includegraphics[page=20,width=\textwidth]{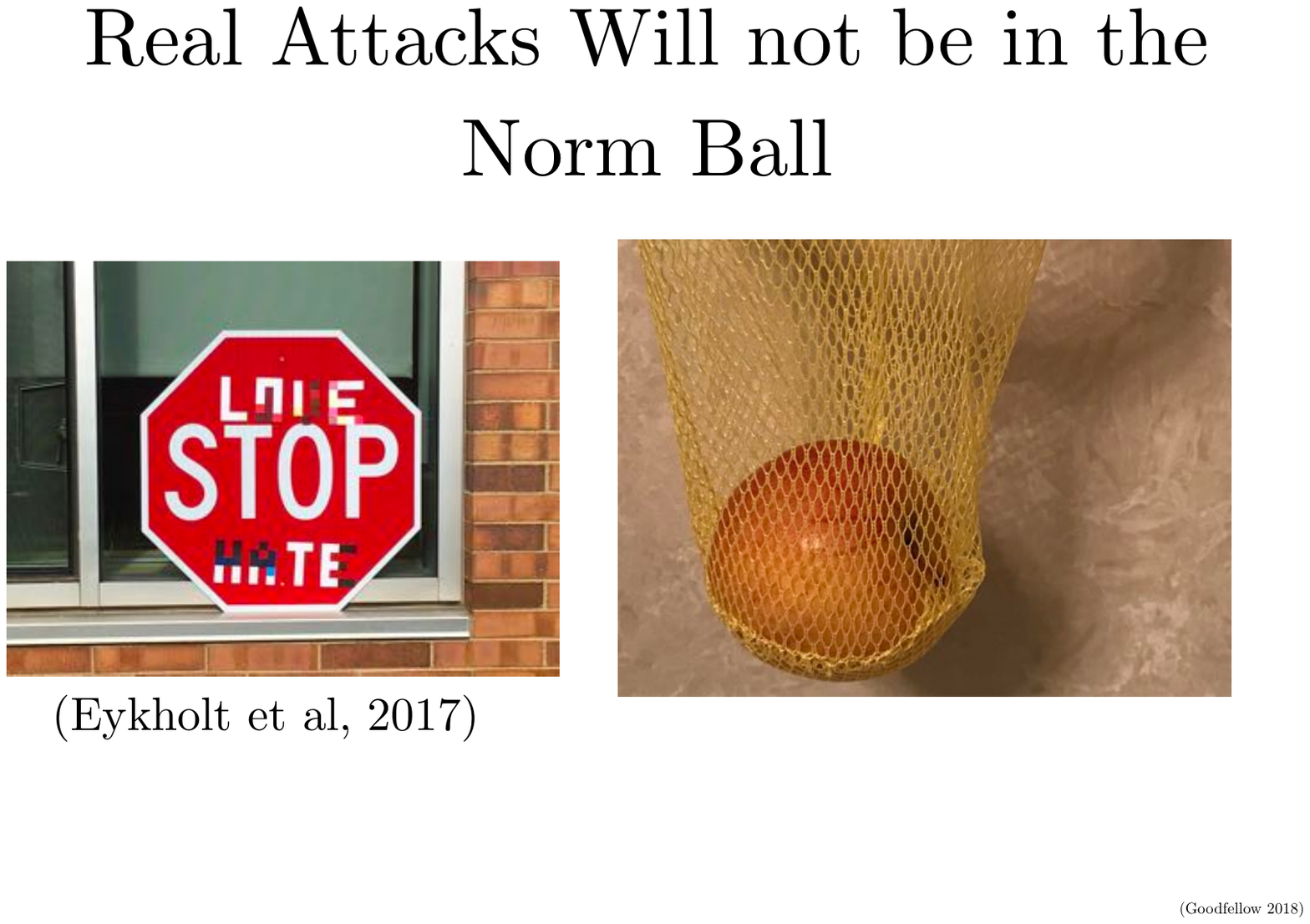}
            
            \caption[20]{
The norm ball is a nice way of formalizing games for basic research purposes. We must remember though that the norm ball is not the real game.}
            
    \end{figure}
    
    \begin{figure}
    \includegraphics[page=21,width=\textwidth]{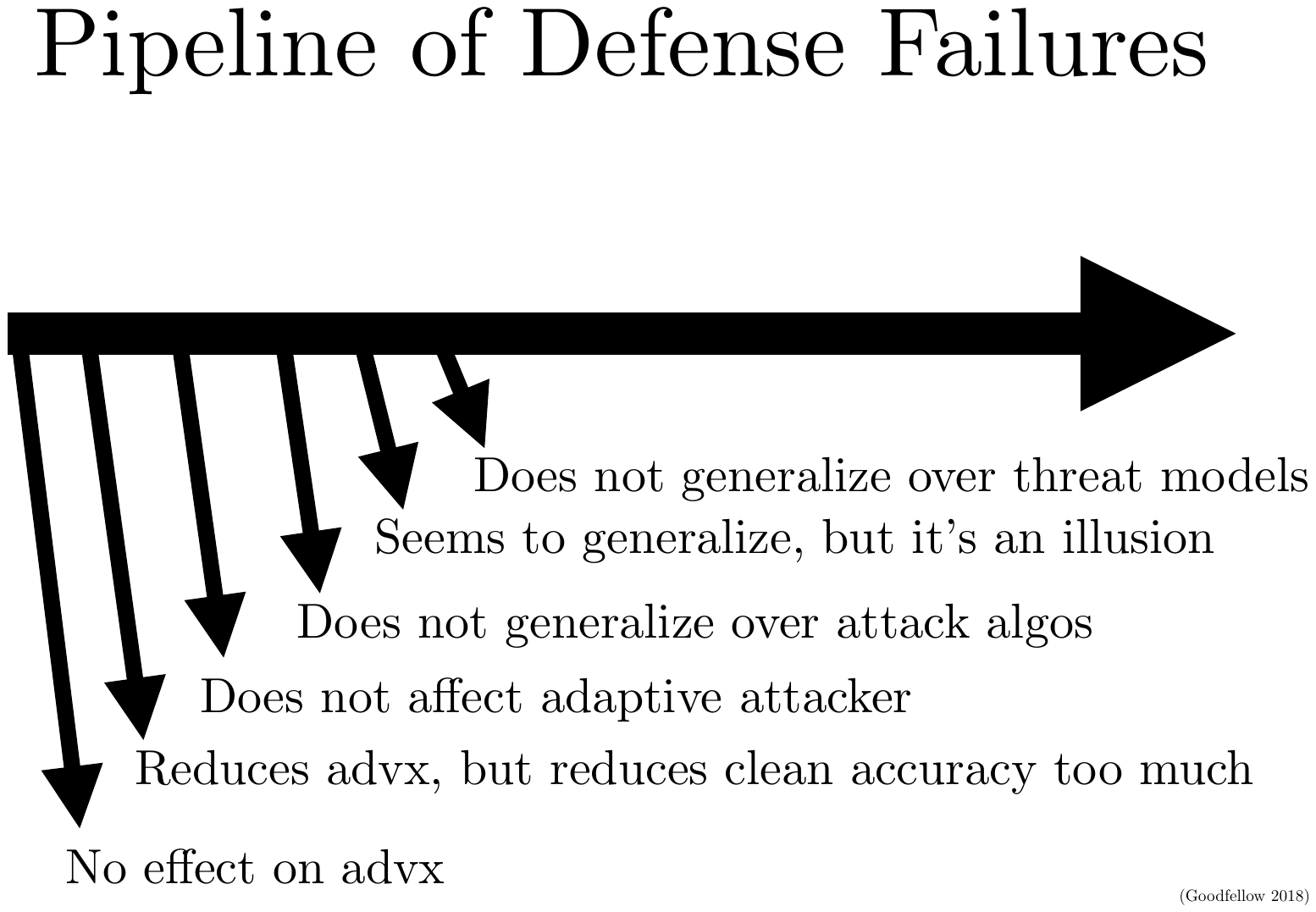}
            
            \caption[21]{}
            
    \end{figure}
    
    \begin{figure}
    \includegraphics[page=22,width=\textwidth]{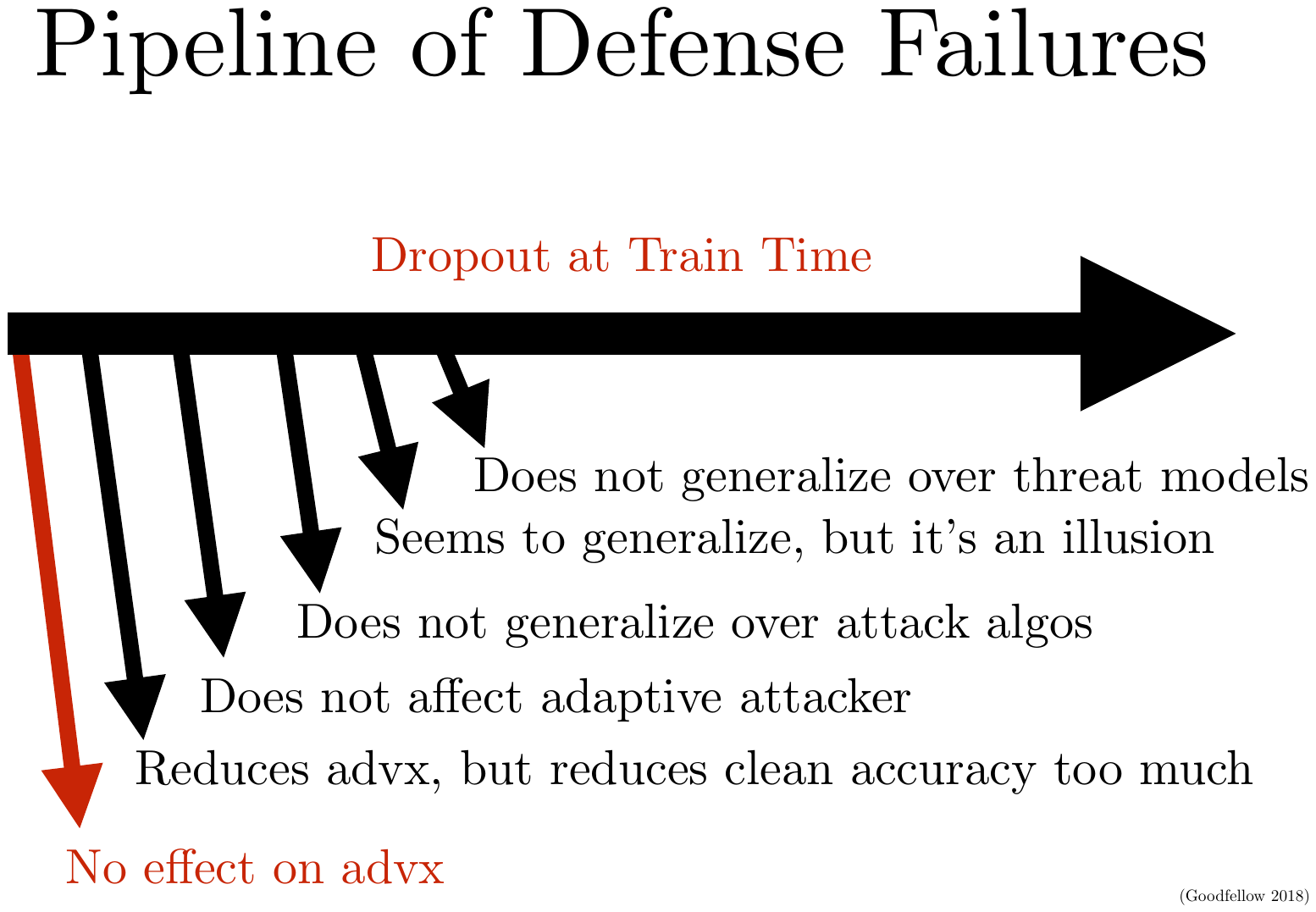}
            
            \caption[22]{
Many of the networks tested in early work on adversarial examples were trained with dropoout: https://arxiv.org/abs/1312.6199 https://arxiv.org/abs/1412.6572

Dropout is a good regularizer, but does not seem to offer any adversarial robustness.}
            
    \end{figure}
    
    \begin{figure}
    \includegraphics[page=23,width=\textwidth]{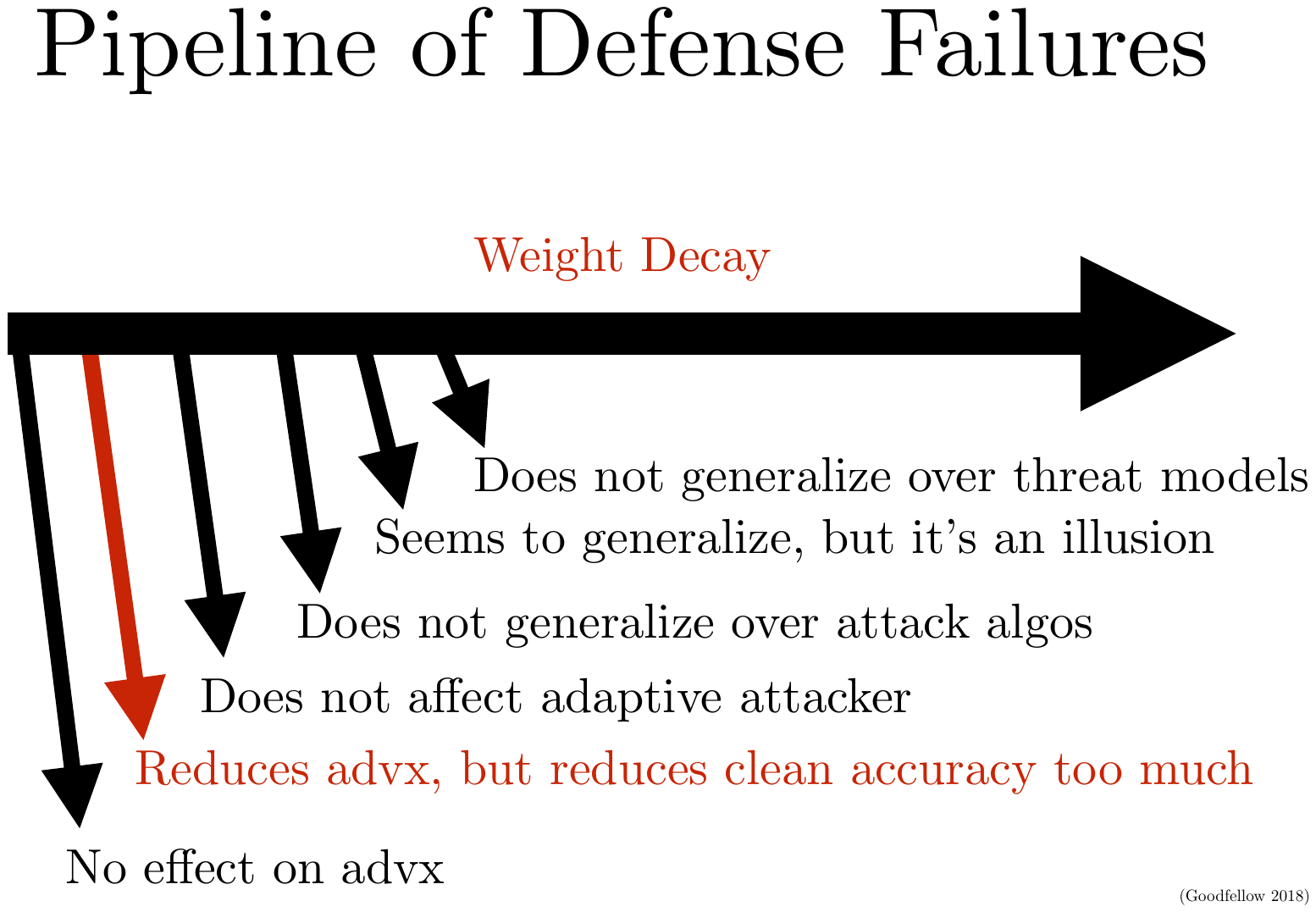}
            
            \caption[23]{
Early work on adversarial examples explored the effect of both L1 and squared L2 weight decay: https://arxiv.org/abs/1312.6199 https://arxiv.org/abs/1412.6572

For large enough weight decay coefficients weight decay does eventually make the weights small enough that the model becomes robust to adversarial examples, but it also makes the accuracy on clean data become relatively bad. Such claims are of course problem-dependent.}
            
    \end{figure}
    
    \begin{figure}
    \includegraphics[page=24,width=\textwidth]{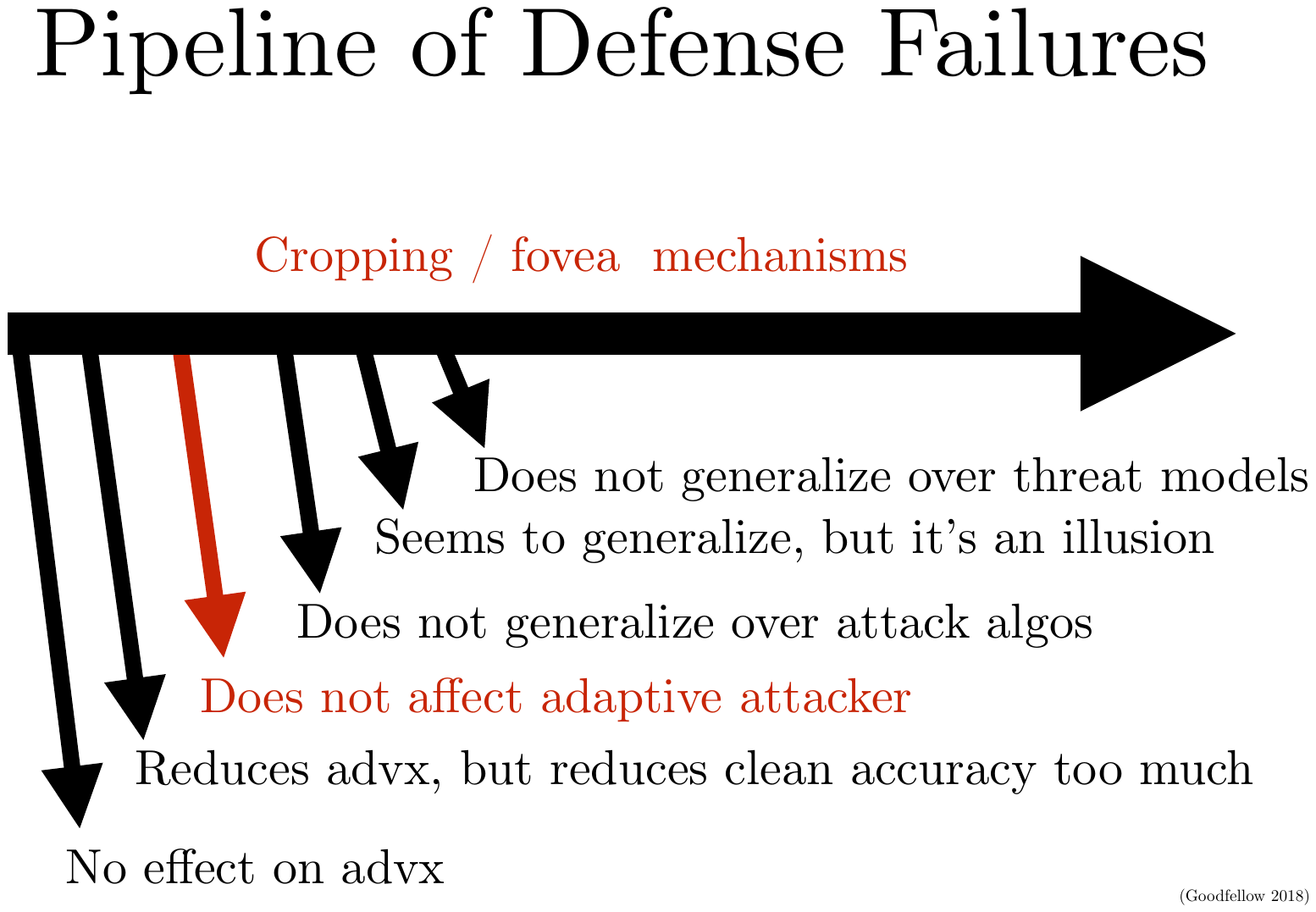}
            
            \caption[24]{
Cropping and fovea mechanisms have been repeatedly proposed as defenses. Generating adversarial examples and then cropping them sometimes reduces error rate. The latest evaluations show that an attacker aware of the mechanism can defeat it. https://arxiv.org/pdf/1802.00420.pdf
}
            
    \end{figure}
    
    \begin{figure}
    \includegraphics[page=25,width=\textwidth]{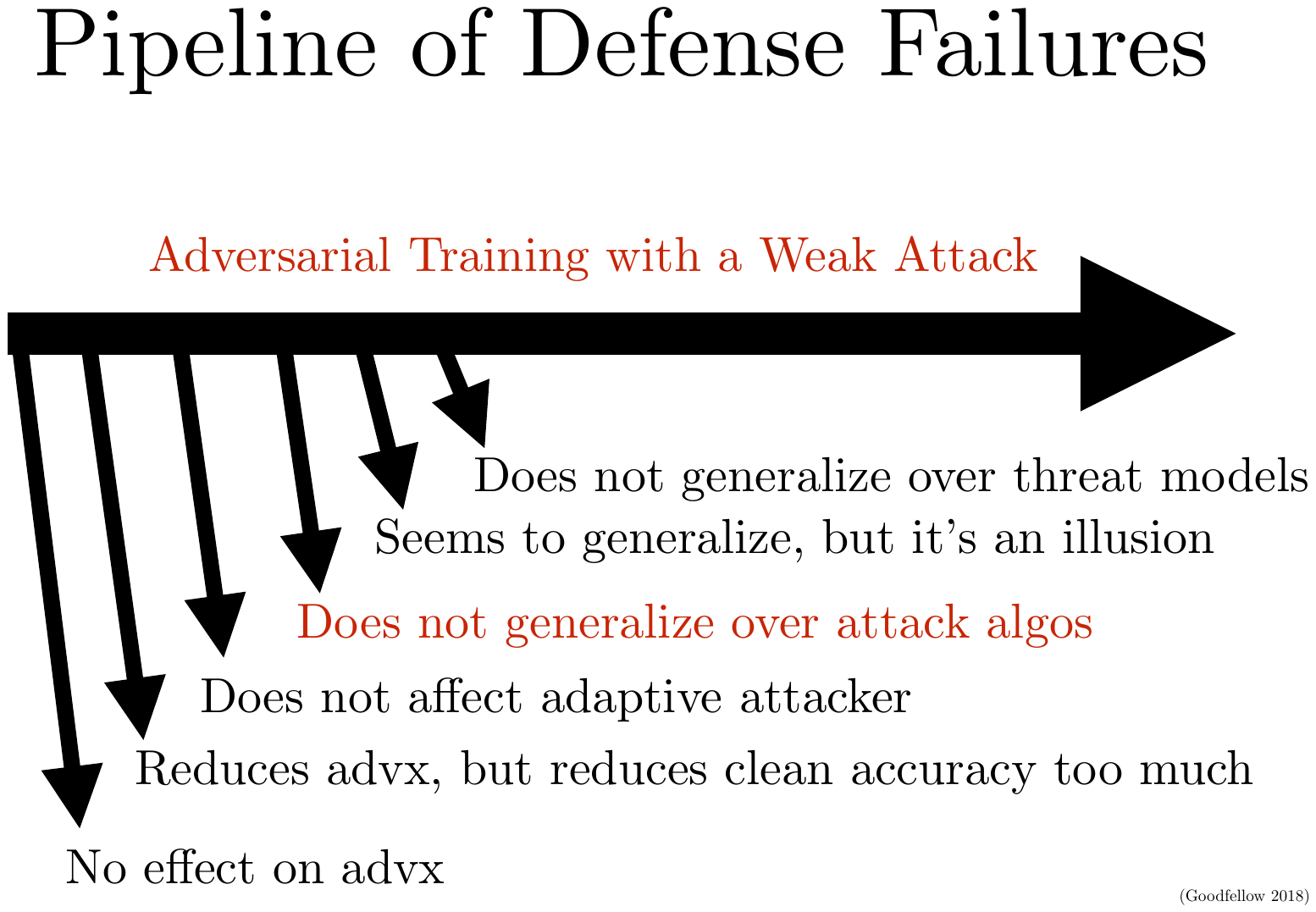}
            
            \caption[25]{
One of the first successes in the defense literature was adversarial training, approximating the minimax optimization using a fast approximation for the adversarial example construction process to generate adversarial examples on the fly in the inner loop of training. This resulted in a model that was robust to adversarial examples made using the same attack algorithm but could be broken by other attack algorithms that used more computation. https://arxiv.org/abs/1412.6572
}
            
    \end{figure}
    
    \begin{figure}
    \includegraphics[page=26,width=\textwidth]{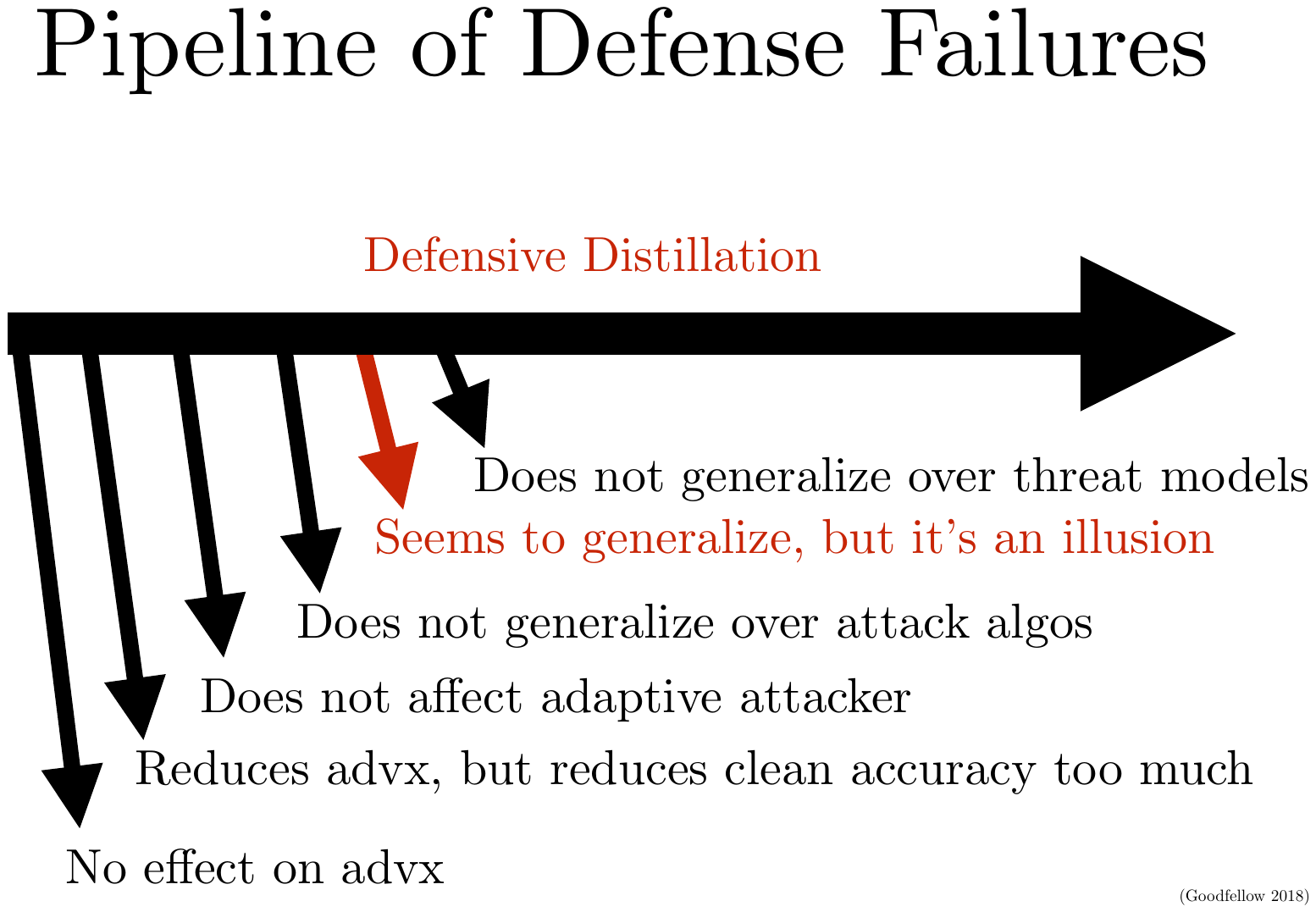}
            
            \caption[26]{
Many defense algorithms seem to perform well against multiple adaptive attack algorithms, but then are later broken. This usually means that their apparent success was an illusion, for example due to gradient masking. In many cases, such broken defenses are still useful contributions to the literature, because they help to develop the stronger attacks that are used to reveal the illusion.

https://arxiv.org/abs/1511.04508
https://arxiv.org/abs/1607.04311
}
            
    \end{figure}
    
    \begin{figure}
    \includegraphics[page=27,width=\textwidth]{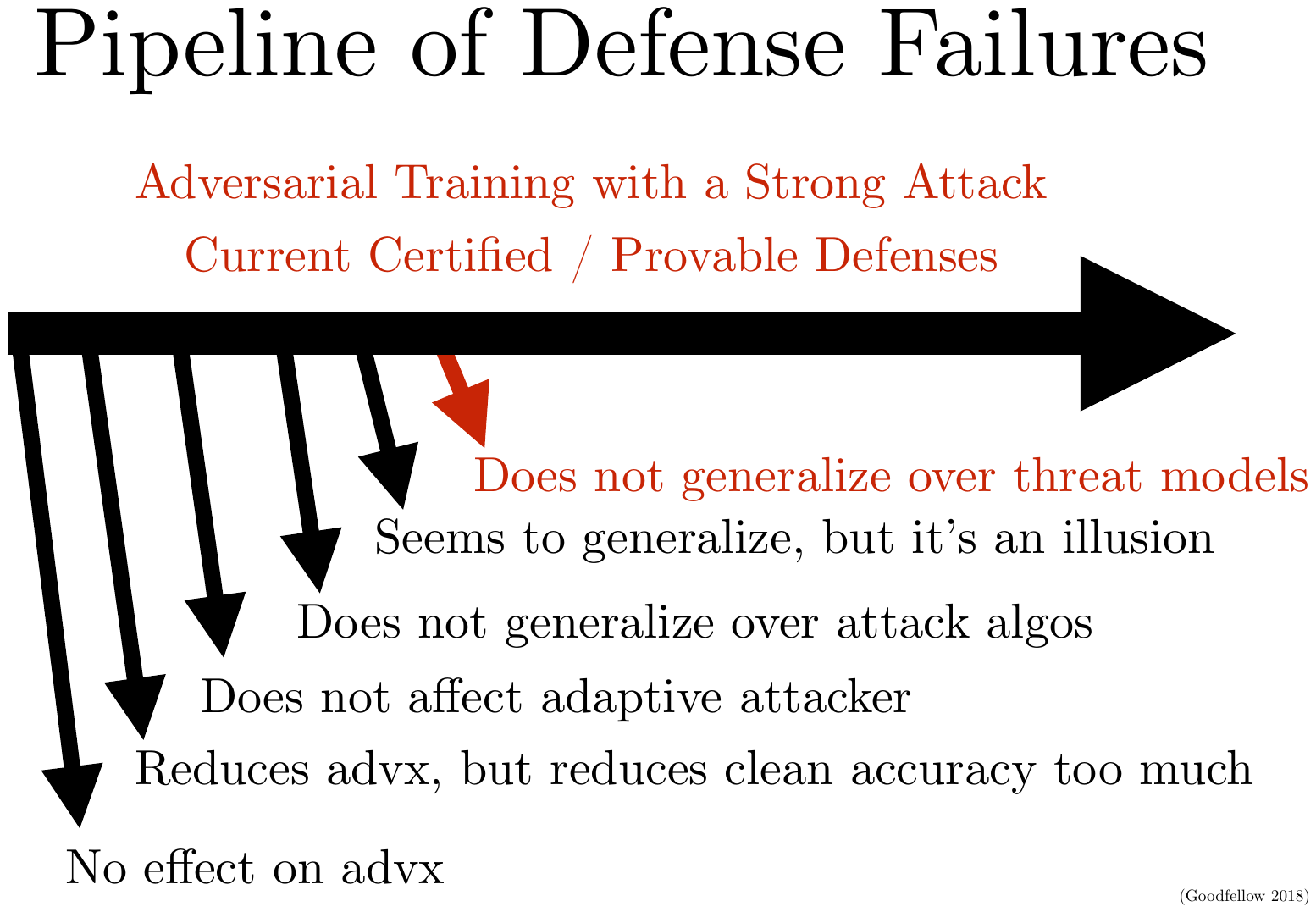}
            
            \caption[27]{
The current state of the art defense ( https://arxiv.org/abs/1803.06373 ) is based on using a strong attack ( https://arxiv.org/abs/1706.06083 ) for adversarial training ( https://arxiv.org/abs/1412.6572 https://arxiv.org/abs/1611.01236 ).

On MNIST in particular, Madry et al's model is regarded as highly robust, after being subject to public scrutiny for several months.

However, this robustness holds only within the $L^\infty$ ball. It is possible to break this model by switching to a different threat model, even one that seems conceptually similar, such as the L1 ball: https://arxiv.org/abs/1709.04114

This problem even applies to all existing certified defenses, because the certificates are specific to a particular norm ball: https://arxiv.org/abs/1801.09344 https://arxiv.org/abs/1803.06567 https://arxiv.org/abs/1711.00851

A late-breaking result is that GAN-based models can also produce adversarial examples that appear unperturbed to a human observer but break these models: https://arxiv.org/pdf/1805.07894.pdf
}
            
    \end{figure}
    
    \begin{figure}
    \includegraphics[page=28,width=\textwidth]{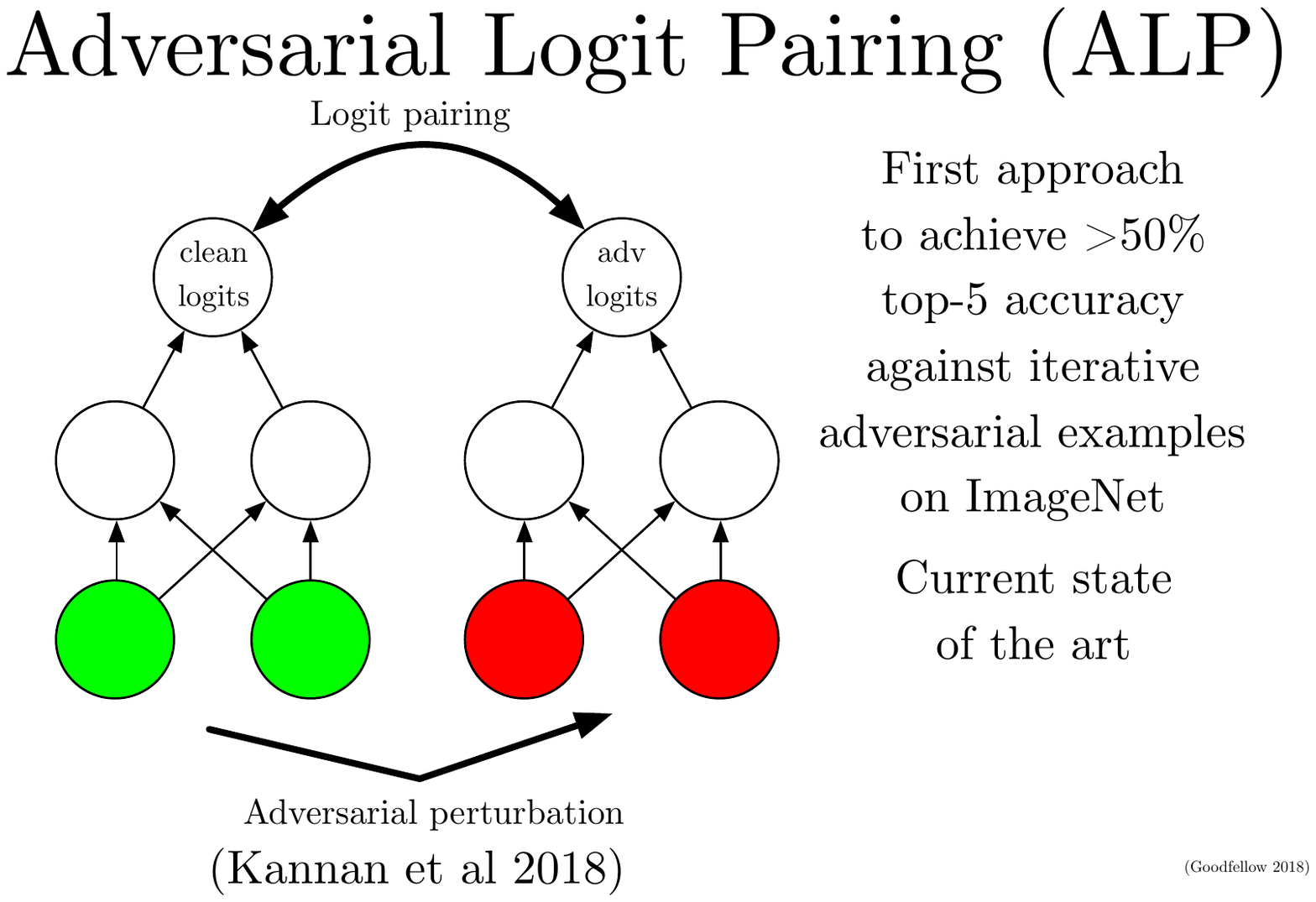}
            
            \caption[28]{
For more information, see https://arxiv.org/abs/1803.06373
}
            
    \end{figure}
    
    \begin{figure}
    \includegraphics[page=29,width=\textwidth]{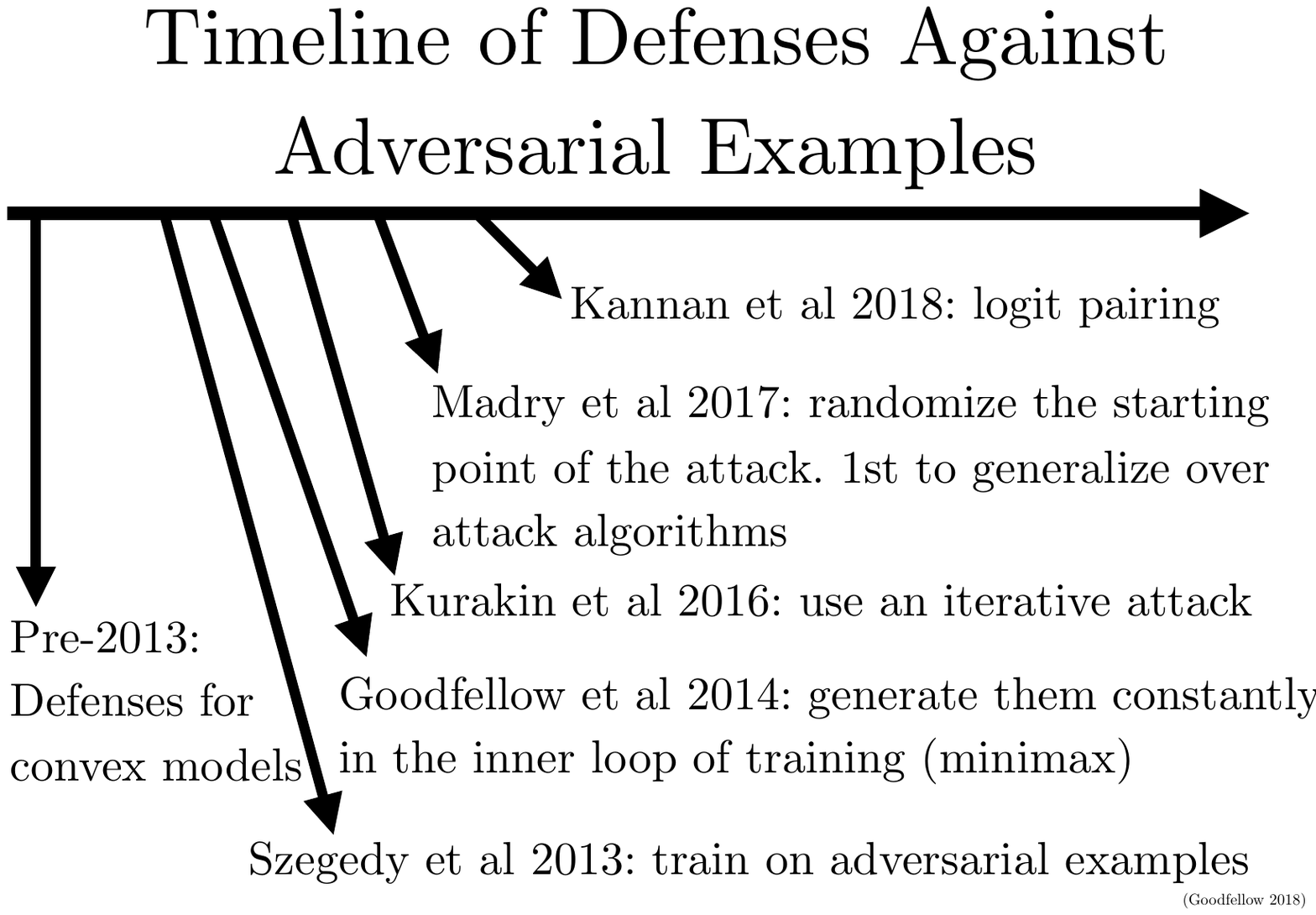}
            
            \caption[29]{
2013: https://arxiv.org/abs/1312.6199
2014: https://arxiv.org/abs/1412.6572
2016: https://arxiv.org/abs/1611.01236
2017: https://arxiv.org/abs/1706.06083
2018: https://arxiv.org/abs/1803.06373

( There has also been earlier work on securing convex models, which is very different from securing neural nets, e.g.: https://homes.cs.washington.edu/~pedrod/papers/kdd04.pdf https://cs.nyu.edu/~roweis/papers/robust\_icml06.pdf )
}
            
    \end{figure}
    
    \begin{figure}
    \includegraphics[page=30,width=\textwidth]{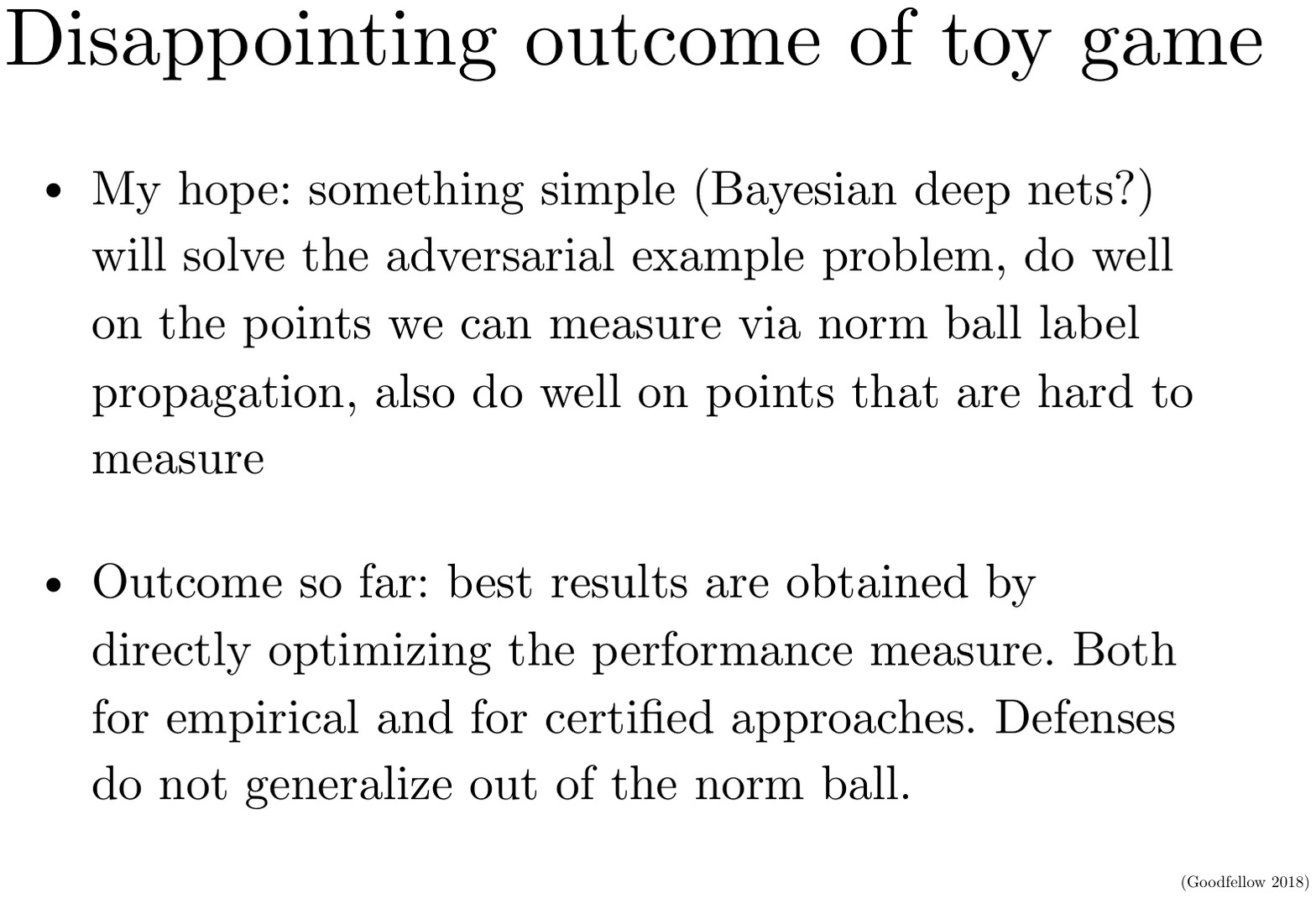}
            
            \caption[30]{}
            
    \end{figure}
    
    \begin{figure}
    \includegraphics[page=31,width=\textwidth]{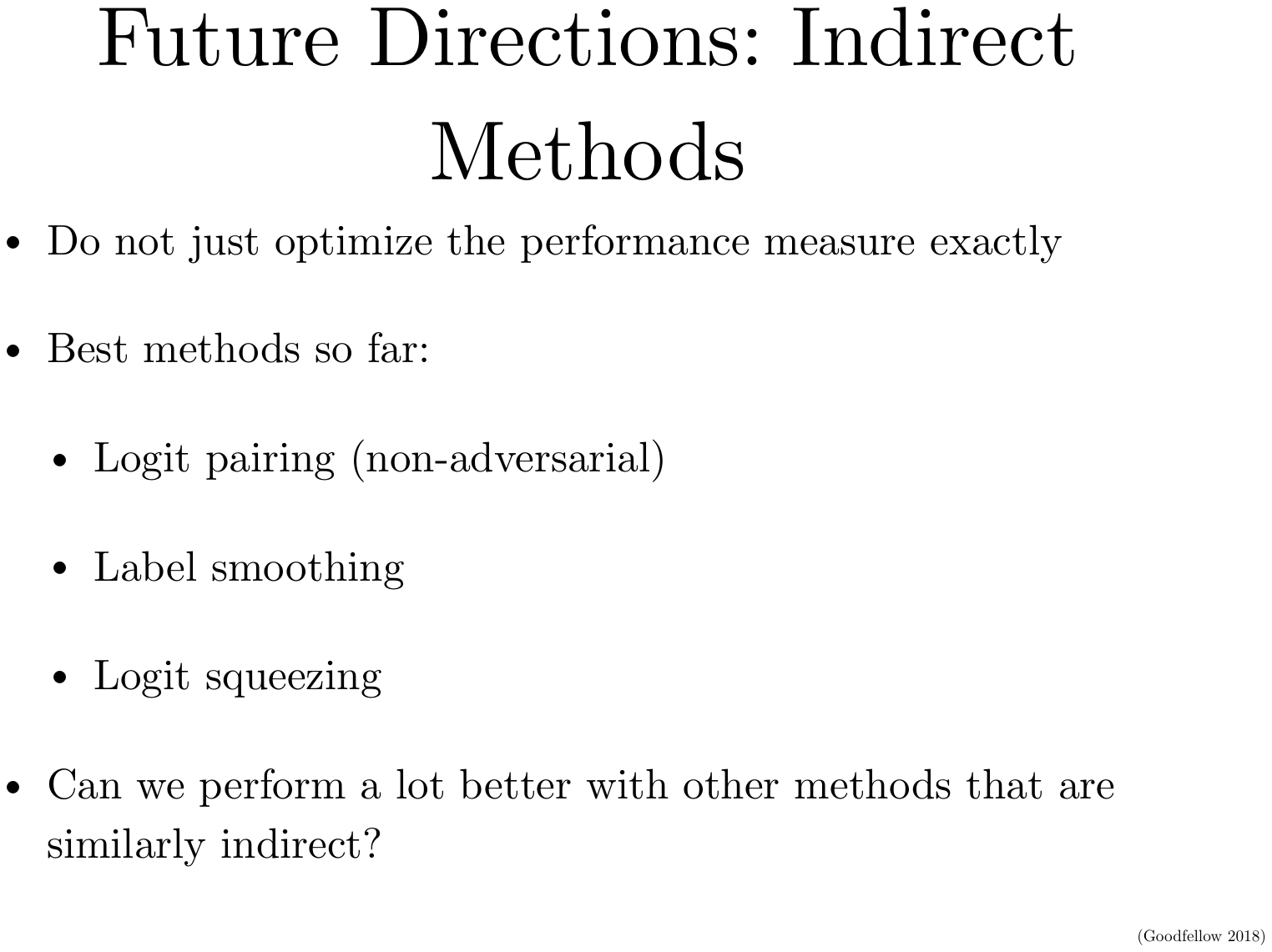}
            
            \caption[31]{
https://books.google.com/books?hl=en\&lr=\&id=F4c3DwAAQBAJ\&oi=fnd\&pg=PA311\&dq=info:J1EtPob5tcoJ:scholar.google.com\&ots=idBTNGuyP8\&sig=HMrjcmrn\_fs2-kyaAo-XNr1Expo\#v=onepage\&q\&f=false observed that label smoothing helps resist adversarial examples.

https://arxiv.org/abs/1803.06373 observed that a few non-adversarial methods of regularizing the logits of a model help it to resist adversarial examples.

So far, these are the only methods of resisting adversarial examples I know of that are not based on directly optimizing some definition of adversarial error rate. I think an important research direction is to find other methods that are similarly indirect and yet perform well. These methods seem most likely to generalize beyond a specific attack model, because they do not involve optimizing directly for performance under that specific attack.
}
            
    \end{figure}
    
    \begin{figure}
    \includegraphics[page=32,width=\textwidth]{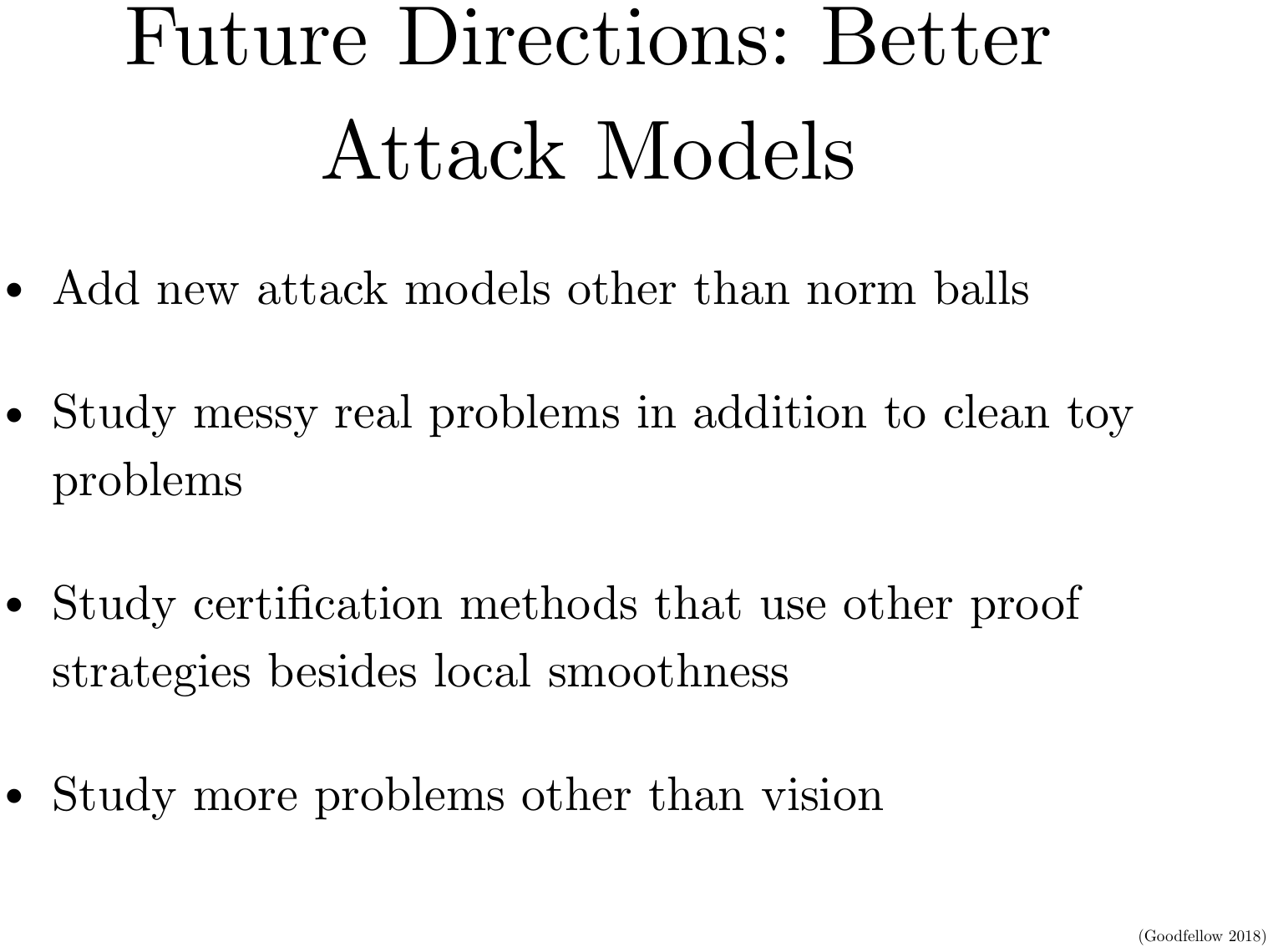}
            
            \caption[32]{}
            
    \end{figure}
    
    \begin{figure}
    \includegraphics[page=33,width=\textwidth]{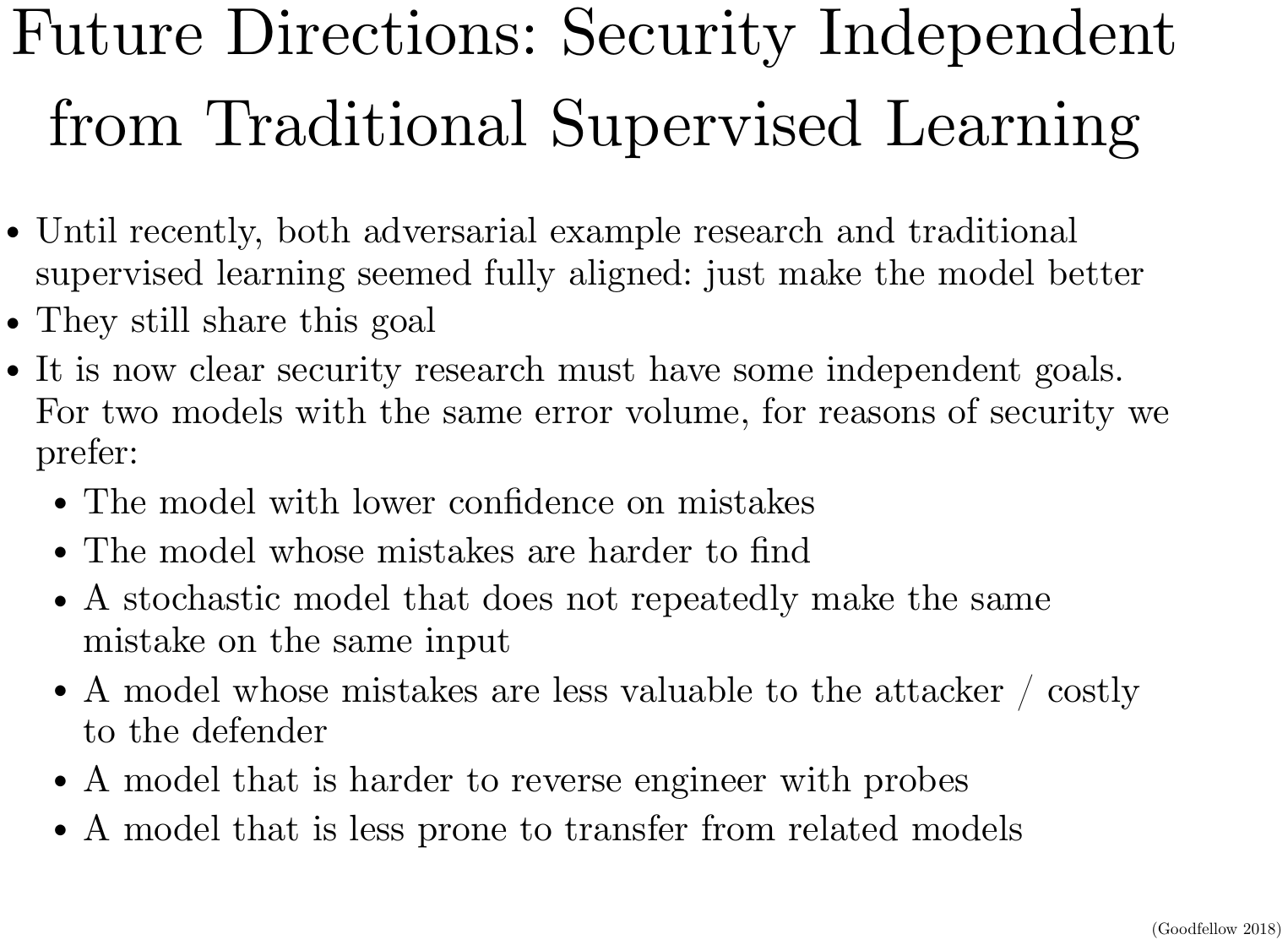}
            
            \caption[33]{}
            
    \end{figure}
    
    \begin{figure}
    \includegraphics[page=34,width=\textwidth]{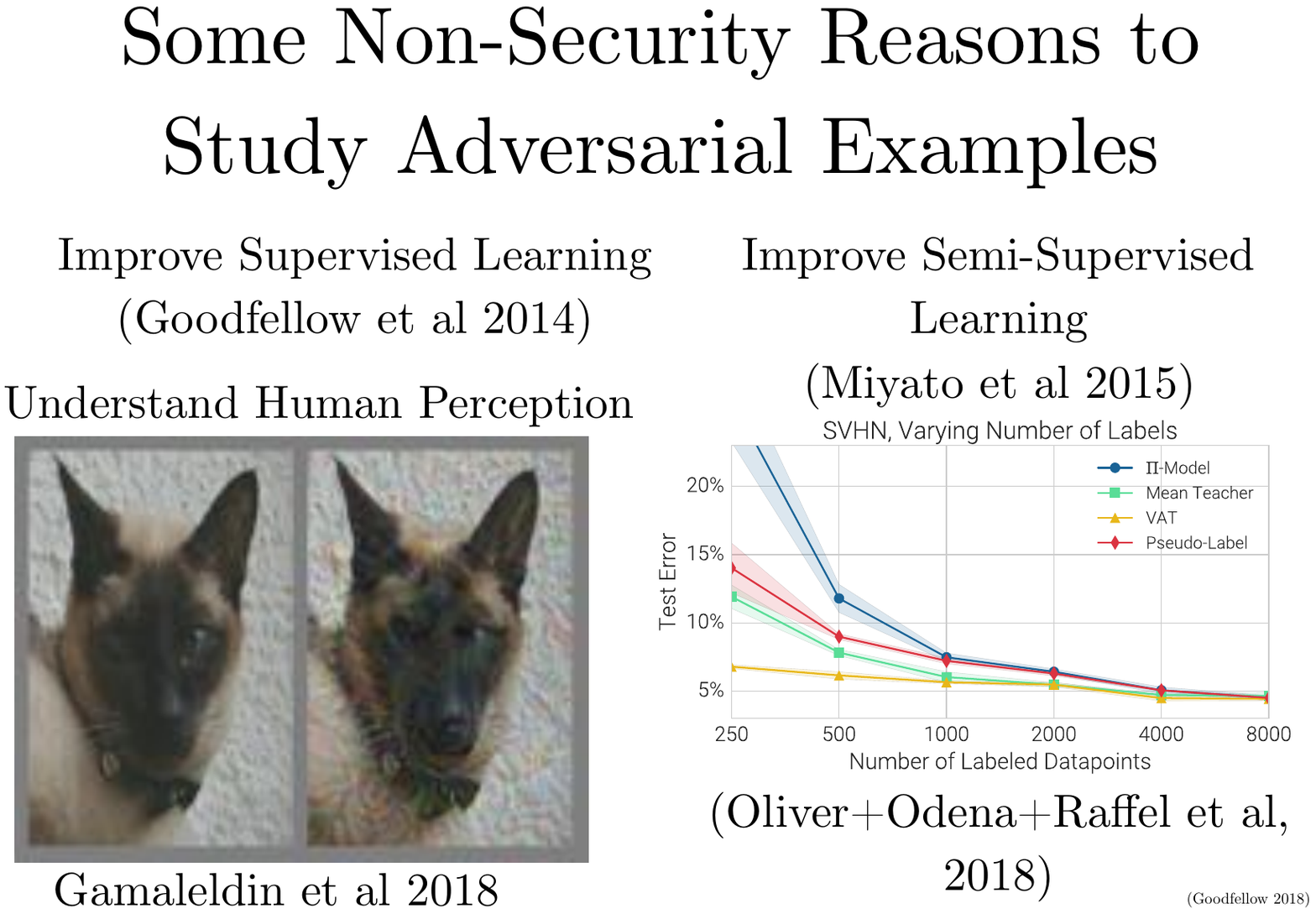}
            
            \caption[34]{
My recommendations today have mostly focused on how to make machine learning secure.

This is distinct from the question of what research should be done regarding adversarial examples.

Studying adversarial examples has improved supervised learning temporarily in the past ( https://arxiv.org/abs/1412.6572 ) and may make a more lasting improvement to supervised learning in the future. Studying adversarial examples has certainly made a significant contribution to semi-supervised learning: virtual adversarial training ( https://arxiv.org/abs/1507.00677 ) was the best performing method in a recent exhaustive benchmark ( https://arxiv.org/abs/1804.09170 ).

Studying adversarial examples may also help to understand the human brain ( https://arxiv.org/abs/1802.08195 ).

Besides these applications, many other applications such as model-based optimization seem possible.
}
            
    \end{figure}
    
    \begin{figure}
    \includegraphics[page=35,width=\textwidth]{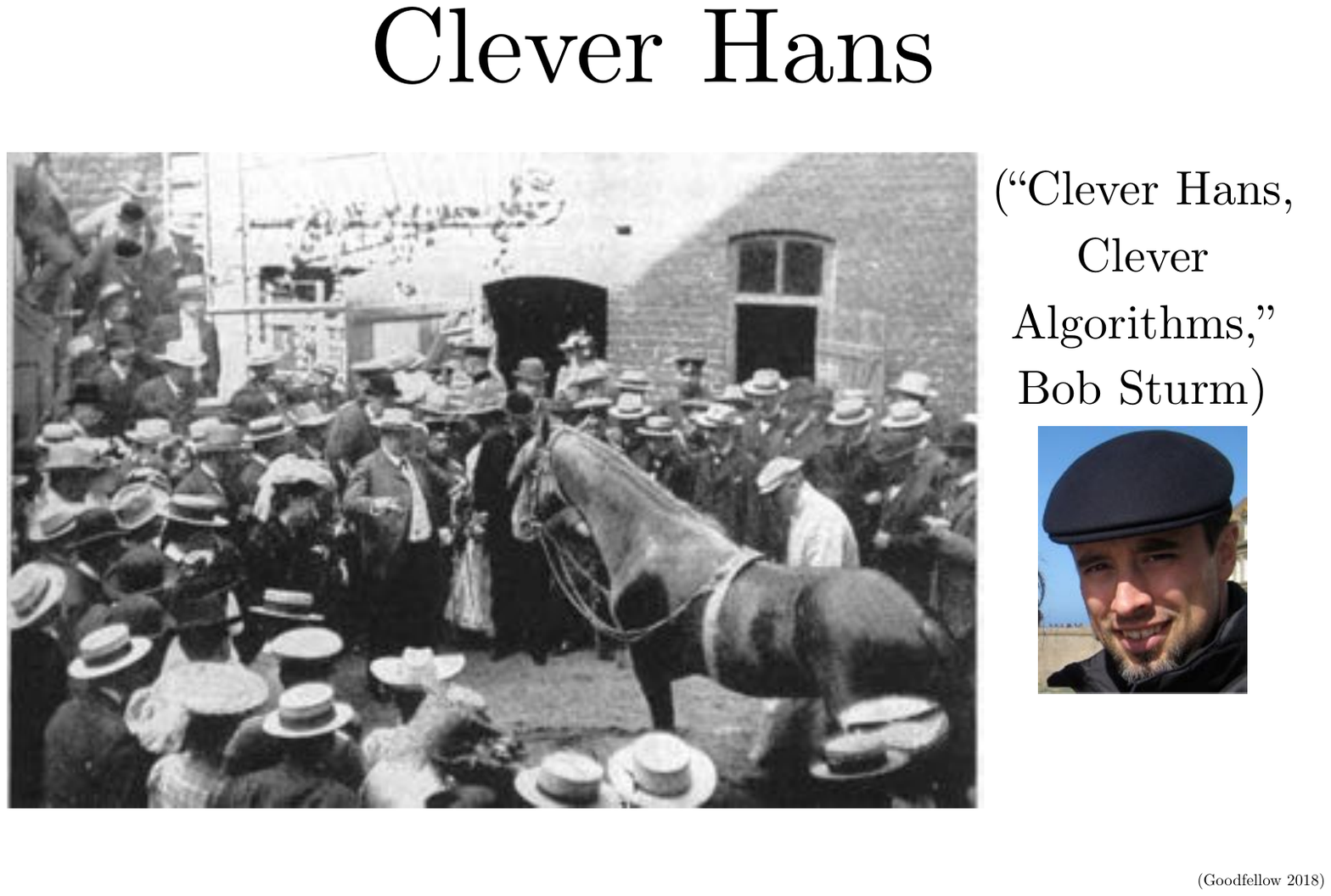}
            
            \caption[35]{
Many adversarial example researchers draw inspiration from the story
of Clever Hans, a horse who was taught to do arithmetic.
Clever Hans appeared able to successfully answer arithmetic questions,
but had actually instead learned to read the questioner's body language
to infer the answer.
Adversarial examples are similar; they show that our algorithms have
learned to apparently succeed at complex tasks, but the algorithms
often get the right answer for the wrong reason.
Studying adversarial examples can help us learn to make algorithms that
actually understand the tasks we assign them to do.
}
            
    \end{figure}
    
    \begin{figure}
    \includegraphics[page=36,width=\textwidth]{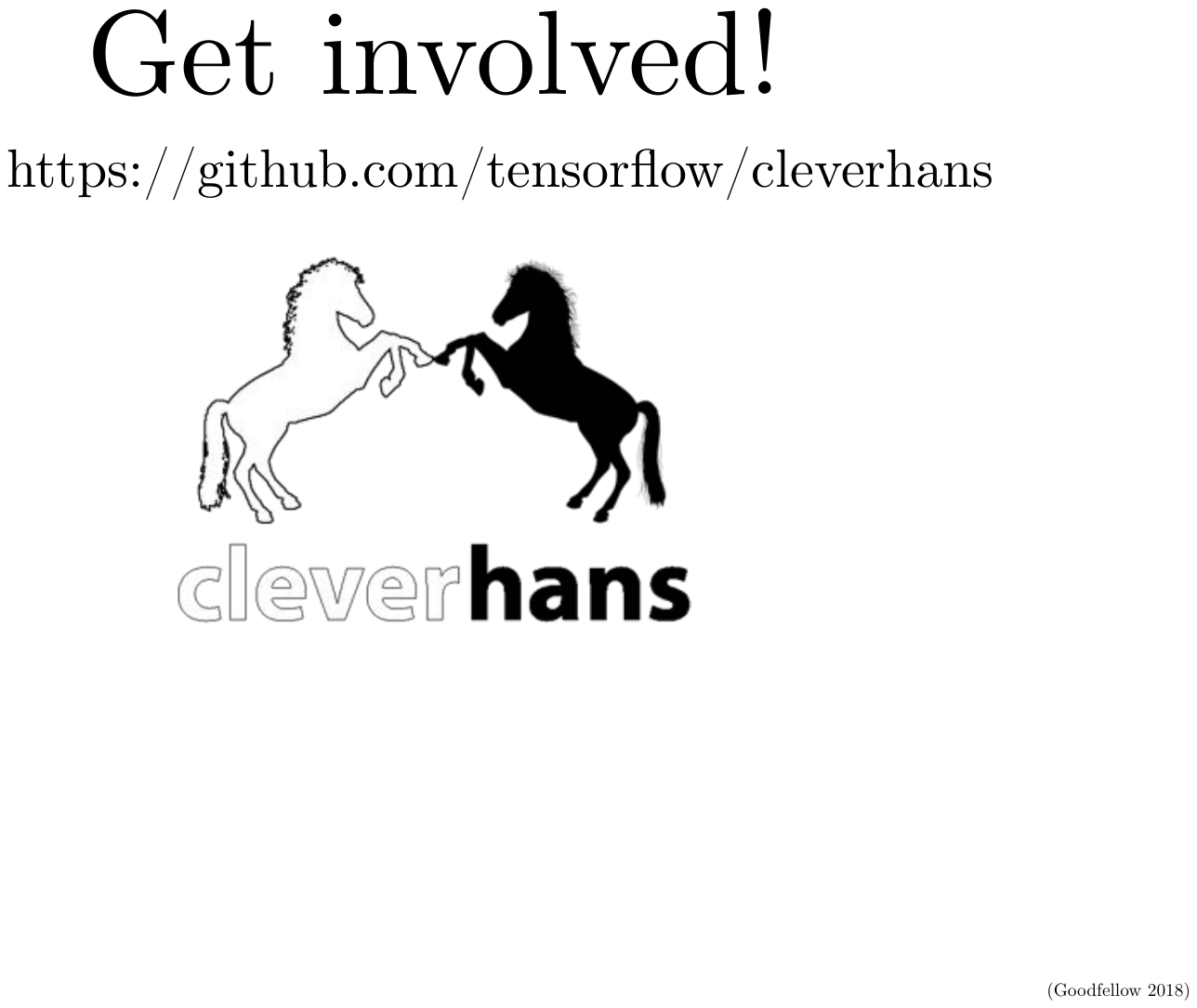}
            
            \caption[36]{
If you would like to get involved in defenses against adversarial examples,
check out the CleverHans library providing reference implementations of
many of the strongest attacks. By benchmarking with this library, you can
be sure that your model is robust to the best known attacks.}
            
    \end{figure}
    
\end{document}